\documentclass[12pt]{article}
\usepackage[authoryear]{natbib}
\usepackage{iclr2023_conference,times}

\usepackage{titlesec}
\titleformat*{\section}{\normalsize\bfseries}
\titleformat*{\subsection}{\normalsize\bfseries}
\titleformat*{\subsubsection}{\normalsize\bfseries}
\usepackage{setspace}
\usepackage{graphicx}

\usepackage{amssymb}
\usepackage{amsmath}
\usepackage{mathtools}
\usepackage{amsthm}
\usepackage{bm}
\usepackage{csquotes}
\usepackage{multirow}
\usepackage{booktabs}
\usepackage{tabularx}
\usepackage{siunitx}
\usepackage{placeins}
\usepackage{url}
\usepackage{hyperref}
\usepackage{algorithmic}
\usepackage{algorithm}
\usepackage{rotating}
 
\theoremstyle{plain}
\newtheorem{theorem}{Theorem}[section]
\newtheorem{proposition}[theorem]{Proposition}

\DeclarePairedDelimiter\autobracket{(}{)}
\newcommand{\br}[1]{\autobracket*{#1}}
\DeclarePairedDelimiter\sparen{[}{]}
\newcommand{\sbr}[1]{\sparen*{#1}}

\DeclareMathOperator*{\argmin}{arg\,min}
\DeclareMathOperator*{\argmax}{arg\,max}

\newcommand{\KL}{D_{\mathrm{KL}}}
\newcommand{\brt}{\br{\bth}}
\newcommand{\natgrad}{\tilde{\nabla}}
\newcommand{\I}{\mathcal{I}_\bz}

\newcommand{\E}{\mathbb{E}}

\renewcommand{\S}{\Sigma}
\newcommand{\iS}{\Sigma^{-1}}
\newcommand{\N}{\mathcal{N}}
\newcommand{\qz}{q_\bz}
\newcommand{\R}{\mathbb{R}}

\newcommand{\iSz}{\Sigma^{-1}_0}
\newcommand{\diag}[1]{\text{diag}\br{#1}}
\newcommand{\s}{\bm{\sigma}^2}
\newcommand{\is}{\bm{\sigma}^{-2}}
\newcommand{\Cov}{\mathrm{Cov}}
\newcommand{\Var}{\mathrm{Var}}
\newcommand{\lik}{p\br{\by \vert \bth}}

\newcommand{\bth}{{\bm{\theta}}}

\newcommand{\bxi}{{\bm{\xi}}}
\newcommand{\bmu}{{\bm{\mu}}}
\newcommand{\bnu}{{\bm{\nu}}}
\newcommand{\by}{{\bm{y}}}

\renewcommand{\vec}[1]{\text{vec}\br{#1}}

\newcommand{\bmz}{{\bm{\zeta}}}

\newcommand{\bz}{{\bm{\zeta^c}}}
\newcommand{\bzp}{{\bm{\zeta^i}}}
\newcommand{\LB}{\mathcal{L}}
\newcommand{\LBz}{\LB\br{\bz}}
\newcommand{\LBzp}{\LB\br{\bzp}}
\newcommand{\lb}{\LB\br{\bmz}}

\iclrfinalcopy
\title{Manifold Gaussian Variational Bayes on the Precision Matrix}
\author{Martin Magris, Mostafa Shabani \& Alexandros Iosifidis \\
Department of Electrical and Computer Engineering\\
Aarhus University\\
Finlandsgade 22, 8200 Aarhus N, Denmark \\
\texttt{\{magris,mshabani,ai\}@ece.au.dk} \\
}

\begin{document}


\maketitle

{\bf \normalsize Martin Magris, Mostafa Shabani, Alexandros Iosifidis}\\
{Department of Electrical and Computer Engineering, Aarhus University, Finlandsgade 22, Aarhus 8200, Denmark}
\\


\thispagestyle{empty}
\markboth{}{NC instructions}

\ \vspace{-0mm}\\

\begin{center} {\bf \normalsize  Abstract} \end{center}
We propose an optimization algorithm for Variational Inference (VI) in complex models. Our approach relies on natural gradient updates where the variational space is a Riemann manifold. We develop an efficient algorithm for Gaussian Variational Inference whose updates satisfy the positive definite constraint on the variational covariance matrix. Our Manifold Gaussian Variational Bayes on the Precision matrix (MGVBP) solution provides simple update rules, is straightforward to implement, and the use of the precision matrix parametrization has a significant computational advantage. Due to its black-box nature, MGVBP stands as a ready-to-use solution for VI in complex models. Over five datasets, we empirically validate our feasible approach on different statistical and econometric models, discussing its performance with respect to baseline methods.

\section{Introduction}\label{sec:intro}

Although Bayesian principles are not new to Machine Learning (ML) \citep{mackay1992bayesian,mackay1995probable,lampinen2001bayesian}, it is only with the recent methodological developments that we are witnessing a growing use of Bayesian techniques in the field are \citep{zhang2018noisy,trusheim2018boosting,osawa_practical_2019,khan2018fastscalable,khan2018fastyetsimple}. 
In typical ML settings, the applicability of sampling methods for the challenging computation of the posterior is prohibitive; however, approximate methods such as Variational Inference (VI) have been proved suitable and successful \citep{saul1996mean,wainwright2008graphical,hoffman2013stochastic,blei2017variational}.
VI is generally performed with Stochastic Gradient Descent (SGD) methods \citep{robbins1951stochastic,hoffman2013stochastic,salimans2014using},
boosted by the use of natural gradients 
\citep{hoffman2013stochastic,wierstra2014natural,khan2018fastscalable}, and the updates often take a simple form \citep{khan2018fastyetsimple,osawa_practical_2019,magris2022quasi}. 

Most VI algorithms rely on the extensive use of models' gradients, and the form of the variational posterior implies additional model-specific derivations that are not easy to adapt to a general, plug-and-play optimizer. 
Black box methods \citep{ranganath2014black} are straightforward to implement and versatile as they avoid model-specific derivations by relying on stochastic sampling \citep{salimans2014using,paisley2012variational,kingma2013auto}. The increased variance in the gradient estimates as opposed to, e.g., methods relying on the reparametrization trick  \citep{blundell2015weight,xu2019variance} can be alleviated with variance reduction techniques.

Furthermore, most existing algorithms do not directly address parameter constraints. Under the typical Gaussian variational assumption, granting positive-definiteness of the covariance matrix is an acknowledged problem \citep{tran_variational_2021,khan2018fastscalable,lin2020handling}. Only a few algorithms directly tackle the problem \citep{osawa_practical_2019,lin2020handling}, see Section \ref{sec:related_works}. A recent approximate approach based on manifold optimization is found in \citep{tran_variational_2021}. For a review of the various algorithms for performing VI, see \citep{magris2022survey}.

On the results of \cite{tran_variational_2021} and on their of Manifold Gaussian Variational Bayes (MGVB) method, we develop a variational inference algorithm that explicitly tackles the positive-definiteness constraint for the variational covariance matrix, resembles the readily-applicable natural-gradient black-box framework of \citep{magris2022quasi}, and that has computational advantages.
We bridge a theoretical issue for the use of symmetric and positive-definite manifold retraction and parallel transport for Gaussian VI, leading to our Manifold Gaussian variational Bayes on the Precision matrix (MGVBP) algorithm.
Our solution, based on the precision matrix parametrization of the variational Gaussian distribution, has furthermore a computational advantage over the implementation of the usual canonical parameterization on the covariance matrix, as the form of the relevant gradients in our update rule is greatly simplified.
We distinguish and apply two forms of the stochastic gradient estimator that are applicable in a wider context and show how to exploit certain forms of the prior/posterior further to reduce the variance of the stochastic gradient estimators. 
We show that MGVBP is straightforward to implement, discuss recommendations and practicalities in this regard, and demonstrate its feasibility in extensive experiments over five datasets, 14 models, three competing VI optimizers, and a Markov Chain Monte Carlo baseline. 

In Section \ref{sec:vi}, we review the basis of VI, in Section \ref{sec:related_works}, we review the Manifold Gaussian Variational Bayes approach and other related works, Section \ref{sec:MGVBP} describes the proposed approach. Section \ref{sec:implementation_aspects} discusses implementation aspects, results are reported in Section \ref{sec:experiments}, and Section \ref{sec:conclusion} concludes this paper. Appendices expand the experiments and provide proofs.

\section{Variational inference}\label{sec:vi}
Variational Inference (VI) is a convenient and feasible approximate method for Bayesian inference. Let $\by$ denote the data, $p\br{\by\vert \bth}$ the likelihood of the data based on some model whose $d$-dimensional parameter is $\bth$. Let $p\brt$ be the prior distribution on $\bth$. In standard Bayesian inference, the posterior is retrieved via the Bayes theorem as $p\br{\bth \vert \by} = p\brt p\br{\by\vert \bth} / p\br{\by}$. As the marginal likelihood $p\br{\by}$ is generally intractable, Bayesian inference is often difficult for complex models. Though sampling techniques can tackle the problem, non-parametric and asymptotically exact Monte Carlo methods may be slow, especially in high-dimensional applications \citep{salimans2015markov}. 

Fixed-form VI approximates the true unknown posterior with a probability density $q$ chosen within a tractable class of distributions $\mathcal{Q}$, such as the exponential family. VI turns the Bayesian inference problem into that of finding the best variational distribution $q^\star \in \mathcal{Q}$ minimizing the Kullback-Leibler (KL) divergence from $q$ to $p\br{\bth \vert \by}$: $q^\star = \argmin_{q \in \mathcal{Q}}  \KL\br{q \vert \vert p\br{\bth\vert \by}}$. It can be shown that the KL minimization problem is equivalent to the maximization of the so-called Lower Bound (LB) on $\log p\br{\by}$, e.g., \citep{tran2021practical}.
The optimization problem accounts for finding the optimal variational parameter $\bmz$ parametrizing $q\equiv q_{\bmz}$ that maximizes the Lower Bound (LB) ($\LB$),
that is $\bmz^\star = \argmax_{\bmz \in \mathcal{Z}} \lb$, with
\begin{align*}
\lb &\coloneqq \int q_\bmz\brt \log \frac{p\brt p\br{y\vert \bth}}{q_\bmz\brt} \mathrm{d} \bth = \E_{q_\bmz}\sbr{\log \frac{p\brt p\br{\by\vert \bth}}{q_\bmz\brt}} = \E_{q_\bmz}\sbr{h_\bmz\brt} \text{,}
\end{align*}
where $\E_{q_\bmz}$ means that the expectation is taken with respect to the distribution $q_\bmz$, and $\mathcal{Z}$ is the parameter space for $\bmz$. 

The maximization of the LB is generally addressed with a gradient-descent method such as SGD \citep{robbins1951stochastic}, ADAM \citep{kingma2014adam}. 
The learning of the parameter $\bmz$ based on standard gradient descent is, however, problematic as it ignores the information geometry of the distribution $q_{\bmz}$, is not scale invariant, unstable, and very susceptible to the initial values \citep{wierstra2014natural}. SGD  implicitly relies on the Euclidean norm for capturing the dissimilarity between two distributions, which can be a poor and misleading measure of discrepancy \citep{khan2018fastyetsimple}. By using the KL divergence in place of the Euclidean norm, the SGD update results in the following natural gradient update: 
\begin{equation}\label{eq:sdg_update}
  \bmz_{t+1} = \bmz_t + \beta_t \left. \sbr{\natgrad_\bmz \lb}\right|_{\bmz = \bmz_t} \text{,}  
\end{equation}
where $\beta_t$ is a possibly adaptive learning rate, and $t$ denotes the iteration.  The above update results in improved steps towards the maximum of the LB when optimizing it for the variational parameter $\bmz$.  The natural gradient $\natgrad_\bmz\lb$ is obtained by rescaling the Euclidean gradient $\nabla_\bmz\lb $ by the inverse of the Fisher Information Matrix (FIM), i.e., 
$$\natgrad_\bmz\lb = \mathcal{I}^{-1}_\bmz \nabla_\bmz\lb \text{,}$$
where $\mathcal{I}_\bmz$ denotes the FIM.
A significant issue in following this approach is that $\bmz$ is unconstrained. Think of a Gaussian variational posterior: in the above setting, there is no guarantee that the covariance matrix updates onto a symmetric and positive definite matrix. As discussed in the introduction, manifold optimization is an attractive possibility.

\section{Elements of manifold optimization}\label{sec:elements_of_manifold}
We wish to optimize the function $\LB$ of the variational parameter $\bm{\zeta}$ 
with an update like \eqref{eq:sdg_update}, where the variational parameter $\bm{\zeta}$ lies in a manifold.
The usual approach for unconstrained optimization reduces to (i) finding the descent direction and (ii) performing a step in that direction to obtain function decrease. The notion of gradient is extended to manifolds through the tangent space.
At a point $\bm{\zeta}$ on the manifold, the tangent space $T_{\bm{\zeta}}$ is the approximating vector space, thus given a descent direction $\bxi_{\bm{\zeta}} \in T_{\bm{\zeta}}$, a step is performed along the smooth curve on the manifold in this direction. 

A Riemannian manifold is a real, smooth manifold equipped with a positive-definite inner product $g_{\bm{\zeta}}\br{\cdot,\cdot}$ on the tangent space at each point $\bm{\zeta}$ (see \citep{absil2008optimization} for a rigorous definition). A Riemann manifold, hereafter simply called manifold, is thus a pair $\br{\mathcal{S},g}$, where $\mathcal{S}$ is a certain set, e.g., of certain matrices. 
For Riemannian manifolds, the Riemann gradient denoted by $\text{grad} f\br{\bm{\zeta}}$ is defined as a direction on the tangent space, where the inner product of the Riemann gradient and any direction in the tangent space gives the directional derivative of the function,
$$
 < \text{grad} f\br{\bm{\zeta}},{\xi_{\bm{\zeta}}}> = \text{D}f\br{\bm{\zeta}}\sbr{{\eta}}\text{,}
$$
where $\text{D}f\br{\bm{\zeta}}\sbr{\eta}$ denotes the directional derivative of $f$ at $\bm{\zeta}$ in the direction $\eta$. The gradient has the property that the direction of $\text{grad} f\br{\bm{\zeta}}$ is the steepest-ascent direction of $f$ at $\bm{\zeta}$ \citep{absil2008optimization}, that important for the scope of optimization.

For a descent direction on the tangent space, the map that gives the corresponding point on the manifold is called the exponential map. The exponential map $\text{Exp}_{\bm{\zeta}}\br{\bxi_{\bm{\zeta}}}$ thus projects a tangent vector $\xi_{\bm{\zeta}}\in T_{\bm{\zeta}}$  back to the manifold, generalizing
the usual concept $\bm{\zeta}+\bxi_{\bm{\zeta}}$ in Euclidean spaces. In fact, $\text{Exp}_{\bm{\zeta}}\br{\bxi_{\bm{\zeta}}}$ can be thought as the point on the manifold reached by leaving from $\bm{\zeta}$ and moving in the direction $\bxi_{\bm{\zeta}}$ while remaining on the manifold. Therefore, in analogy with the usual gradient descent approach $\zeta \leftarrow \zeta + \beta \nabla f\br{\bm{\zeta}}$ with $\beta$ being the learning rate, on manifolds, the update is performed through retraction following the steepest direction provided by the Riemann gradient as $\text{Exp}_{\bm{\zeta} }\br{\beta \,\text{grad} f\br{\bm{\zeta}}}$.

In practice, exponential maps are cumbersome to compute; retractions are used as first-order approximations. A Riemannian manifold also has a natural way of transporting vectors. Parallel transport moves tangent vectors from one tangent space to another while preserving the original length and direction, extending the use of momentum gradients to manifolds.
As for the exponential map, a parallel transport is in practice approximated by the so-called vector transport. Note that the forms of retraction and vector transport, as much as that of the Riemann gradient, depend on the specific metric adopted in the tangent space. 

Thinking of $\bm{\zeta}$ as the parameter of a Gaussian distribution, $\bm{\zeta}$ involves elements related to $\bm{\mu}$, unconstrained over $\R^d$, and elements related to the covariance matrix, constrained to define a valid covariance matrix: the product space of Riemannian manifolds is itself a Riemannian manifold. The exponential map, gradient, and parallel transport are defined as the Cartesian product of the individual ones, while the inner product is defined as the sum of the inner product of the components in their respective manifolds \citep{hosseini2015matrix}.
\begin{figure}
    \centering
    \includegraphics[trim={0cm 0cm 0cm 0cm},clip,width = 0.3\textwidth]{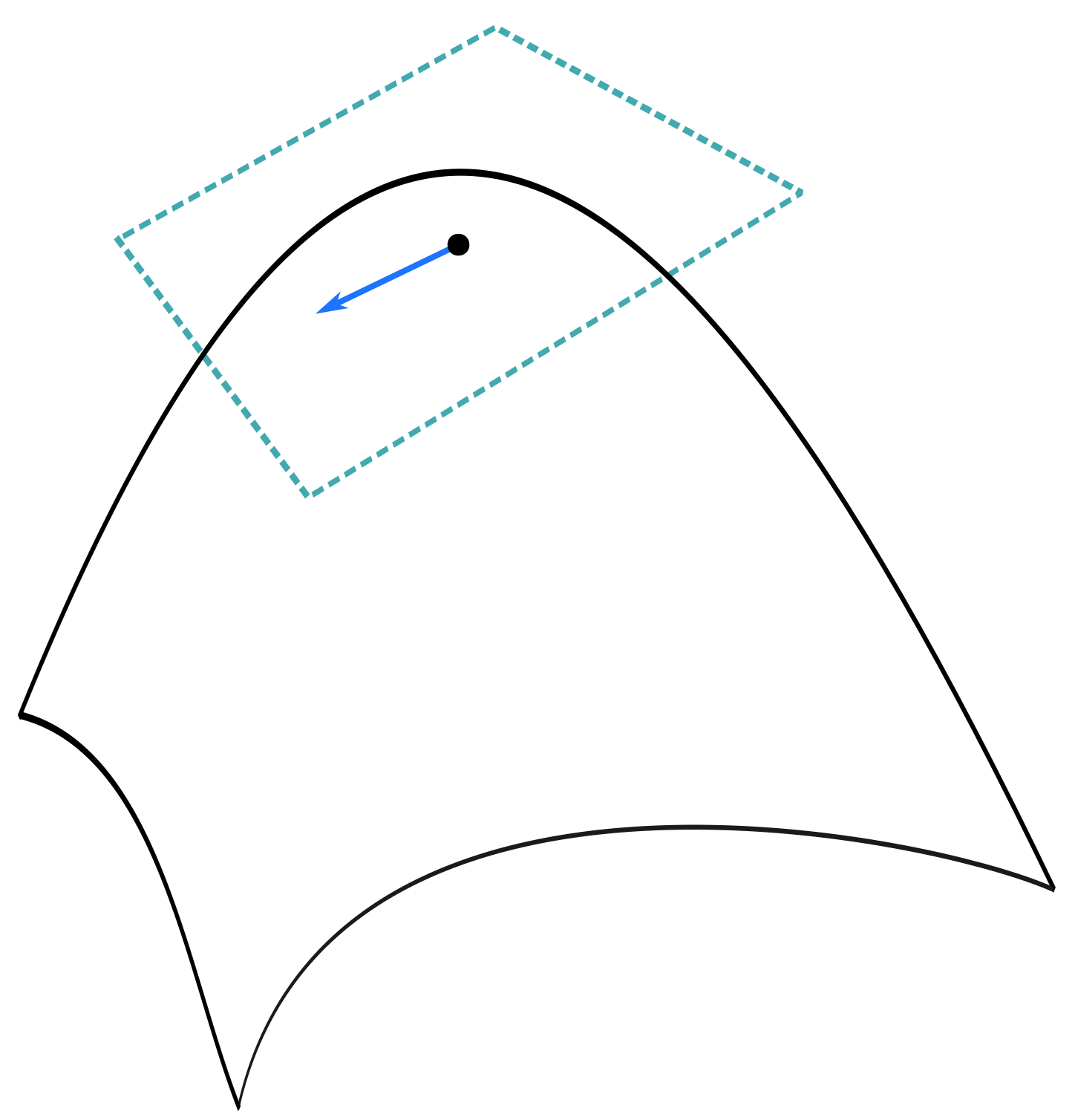}
    \includegraphics[trim={0cm 0cm 0cm 0cm},clip,width = 0.3\textwidth]{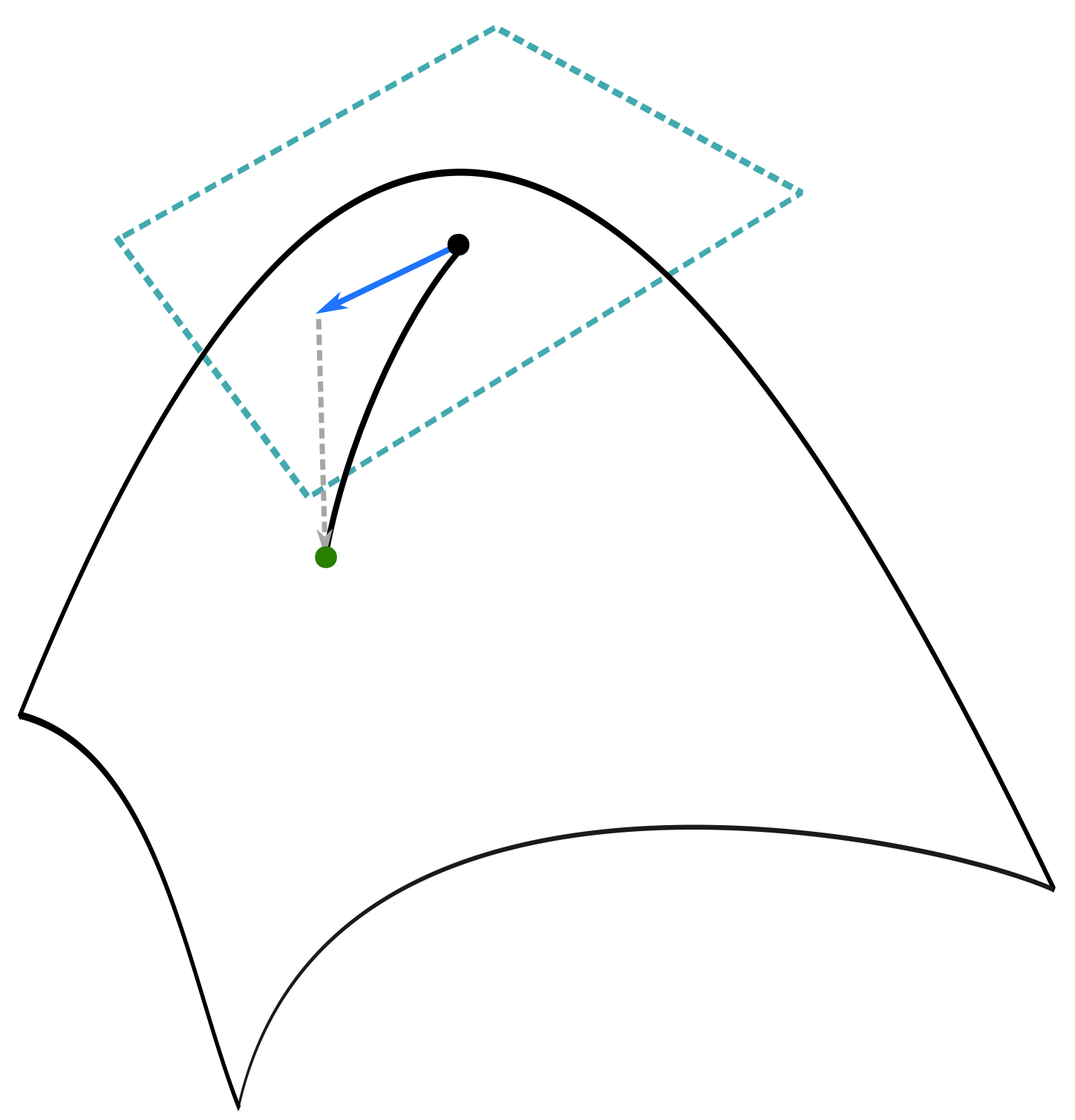}
    \includegraphics[trim={0cm 0cm 0cm 0cm},clip,width = 0.3\textwidth]{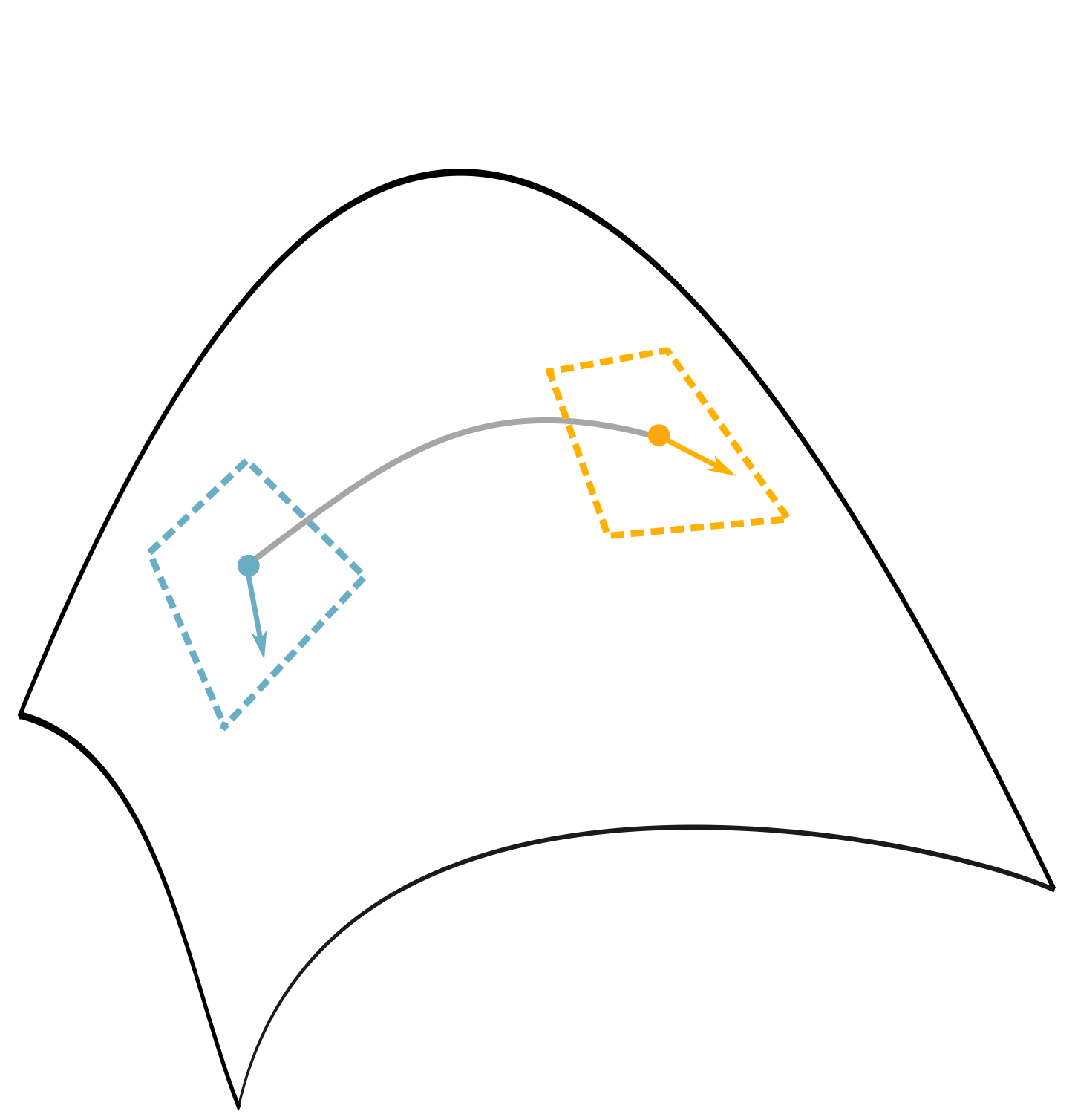}
    \caption{Manifold illustration. Left: manifold (black), tangent space (light blue), and Riemann gradient at the point in black. Middle: exponential map (dotted gray) and the corresponding point on the manifold (green point). Right: Parallel transform between vectors on two tangent planes.}
    \label{fig:manifold}
\end{figure}
\subsection{Two manifolds}\label{subsec:two_manifolds}

Be $\mathcal{S} = \left\lbrace P \in \R^{d\times d}: P = P^\top, P \succ 0 \right\rbrace$ the set of Symmetric and Positive Definite (SPD) $d\times d$ matrices, we denote by $\mathcal{M} =\br{\mathcal{S},g_P}$ the corresponding manifold, where with the metric $g_P$ between two vectors $\bm{\zeta}$ and $\bxi$ at $P$ defined as $g_P = \text{Tr}\br{P^{-1}\bm{\zeta}P^{-1}\bxi}$. In the remainder of the paper, as the metric $g_P$ is derived from the Frobenius norm, i.e., the Euclidean norm of a matrix, we refer to this metric between tangent vectors as Euclidean metric (see, e.g., \citep{bhatia2019bures,han2021riemannian}). By denoting with $f$ a generic smooth function of $P$, from, e.g., \citep{hosseini2015matrix, pennec2020manifold, manopt2014}, the relationship between the usual Euclidean gradient $\nabla f\br{P}$ and the Riemann gradient is 
\begin{equation}\label{eq:riemann_grad_M_main}
\text{grad}\,f\br{P} = P \nabla f\br{P} P \text{,}    
\end{equation} assuming $\nabla f\br{P}$ is  a symmetric matrix.
For the manifold $\mathcal{M}$, the retraction at point $P$ in the direction $\bxi$ is computed as
\begin{equation}\label{eq:retraction_general}
    R_{P}\br{\bxi} = P +\bxi +\frac{1}{2}\bxi P^{-1} \bxi, \qquad\bxi\in T_{P}\mathcal{M}\text{,}
\end{equation}
while the vector transport of the tangent vector from $P_1$ to $P_2$ is given by:
\begin{equation} \label{eq:vectransport_general}
 \mathcal{T}_{P_1\rightarrow P_2}\br{\bxi} = E\bxi E^\top\text{,}\qquad E=\br{P_2 P_1^{-1}}^{\frac{1}{2}}   \text{.}
\end{equation}

In alternative to the manifold $\mathcal{M}$, as for the discussion in Section \ref{sec:vi}, another metric that gained popularity is the Fisher-Rao metric. The Fisher-Rao metric depends on the Fisher Information Matrix (FIM) $\mathcal{I}_P$, and for two vectors  $\bm{\zeta},\bm{\xi}$ at a point $P$ in the tangent space $T_P$, it defines the inner product as $<\bm{\zeta},\bm{\xi}> = \bm{\zeta}^\top \mathcal{I}_P \bm{\xi}$. We refer to the manifold of the SPD matrices equipped with the Fisher-Rao metric as $\mathcal{F}$. Importantly, the manifold $\mathcal{F}$ is the one adopted by the work of \citep{tran_variational_2021}, at the basis of this paper. \citep{tran_variational_2021} furthermore show that the Riemann gradient in $\mathcal{F}$ corresponds to the natural gradient, i.e.,
$$
\text{grad} f \br{P} = \natgrad f\br{P}\text{.}
$$
From the form of the Gaussian FIM, $\natgrad f\br{P} = 2P\nabla f\br{P} P$ (see Appendix \ref{app:proof_prop}), note that 
\begin{equation}\label{eq:analogy}
P \nabla f\br{P} P = \frac{1}{2} \natgrad f\br{P}\text{.}
\end{equation}
I.e., the Riemann gradient in $\mathcal{M}$ is one-half of the natural gradient, which conversely is the Riemann gradient in $\mathcal{F}$. Note that the above also applies to $P^{-1}$ since the inverse of an SPD matrix is as well SPD.

\section{Related work} \label{sec:related_works}

\citep{tran_variational_2021} adopts a fixed-form $d$-dimensional Gaussian distribution with mean $\bmu$ and covariance matrix $\S$ for the variational approximation $q_\bz$, where the variational parameter $\bz$ collects all the elements of $\bmu$ and $\S$, i.e. $\bz = \br{\bmu^\top,\text{vec}\br{\S}^\top}^\top$ (\textit{canonical} parametrization).
There are no restrictions on $\bmu$, yet the covariance matrix $\S$ is constrained to the manifold $\mathcal{F}$ of SPD matrices equipped with the Fisher-Rao metric. 

For a multivariate Gaussian $q_\bz$, the exact form of the corresponding FIM  reads \citep[e.g.,][]{mardia1984maximum}
\begin{equation}\label{eq:gaussian_fim}
\I = \begin{pmatrix}
\iS & 0\\
0 & \I\br{\S}
\end{pmatrix} \text{,}
\end{equation}
with $\I\br{\S}_{\sigma_{ij},\sigma_{kl}} = \frac{1}{2}\text{Tr}\br{\iS \frac{\partial \S}{\partial \sigma_{ij}}\iS \frac{\partial \S}{\partial \sigma_{kl}}}$ being the generic element of the   $d^2 \times d^2$ matrix $\I\br{\S}$.
The Manifold Gaussian Variational Bayes (MGVB) method  \citep{tran_variational_2021} relies on the approximation $ \I\br{\S}\approx \iS \otimes \iS$, where $\otimes$ denotes the Kronecker product.
Accordingly, \citep{tran_variational_2021} compute the convenient approximate form of the natural gradients of the LB with respect to $\bmu$ and $\S$, respectively as\footnote{ \normalsize \doublespacing We follow the presentation of \citep{tran_variational_2021}, where $\natgrad_\S\LBz$ is approximated up to a scaling constant as $\S\nabla_\S\S$. Indeed \citep{lin2020handling} clarify that exact form of $\I\br{\S}$ should read $2(\iS \otimes \iS)$, see e.g., \citep{barfoot2020multivariate}, leading to the exact natural gradient $\natgrad_\S\LBz = 2\S\nabla_\S\LBz\S$.\label{footnote:1}},
\begin{align}\label{eq:grad_S_tran}
    \natgrad_\bmu \LBz = \S \nabla_\bmu\LBz  \quad \text{and} \quad
    \natgrad_\S \LBz \approx  \S \nabla_\S \LBz \S \text{.}
\end{align}
In virtue of the natural gradient definition, the first natural gradient is exact, while the second is approximate.
Thus, \citep{tran_variational_2021} adopts the following updates on the variational parameter:
\begin{equation}\label{eq:tran_updates}
  \bmu \leftarrow \bmu + \beta \natgrad_\bmu \LBz    \quad  \textrm{and}  \quad
  \S \leftarrow R_\S(\beta\natgrad_\S\LBz) \text{,} 
\end{equation}
where $R_\S\br{\cdot}$ denotes the retraction form \eqref{eq:retraction_general}. Retractions are of central importance in manifold optimization and for the scope of the paper: they enable projecting vectors from the so-called tangent space back to the manifold. Momentum gradients can be used in place of plain natural ones. In particular, the momentum gradient for the update of $\S$ relies on a vector transport, granting that, at each iteration, the weighted gradient remains in the tangent space of the manifold, see Section \ref{sec:elements_of_manifold}.

A method of handling the positivity constraint in diagonal covariance matrices is the Variational Online Gauss-Newton (VOGN) optimizer \citep{khan2018fastscalable,osawa_practical_2019}. VOGN relates to the VON \citep{khan2017conjugate} update as it indirectly updates $\bmu$ and $\S$ from Gaussian natural parameters updates. Following a non-Black-Box approach, VOGN uses some theoretical results on the Gaussian distribution to recover an update for $\S$ that involves the Hessian of the likelihood. Such Hessian is estimated as the samples' mean squared gradient, granting the non-negativity of the diagonal covariance update. \citep{osawa_practical_2019} devise the computation of the approximate Hessian in a block-diagonal fashion within the layers of a Deep-learning model. 

\citep{lin2020handling} extend the above to handle the positive definiteness constraint by adding an additional term to the update rule for $\S$, applicable to certain partitioned structures of the FIM. The retraction map in \citep{lin2020handling} is more general than \citep{tran_variational_2021} and obtained through a different Riemann metric, from which MGVB is retrieved as a special case.
As opposed to MGVBP, the use of the reparametrization trick in \citep{lin2020handling} requires model-specific computation or auto-differentiation. See \citep{lin2021structured} for an extension on stochastic, non-convex problems. 

Alternative methods that rely on unconstrained transformations (e.g., Cholesky factor, \citep{tan2021natural}), or on the adaptive adjustment of the learning rate (e.g., \citep{khan2017conjugate}) lie outside the manifold context discussed here. 
Among the methods that do not control for the positive definiteness constraint, the QBVI update \citep{magris2022quasi} provides a comparable black-bock method that, despite other black-box VI algorithms, uses exact natural gradients updates obtained without the computation of the FIM.

\section{Manifold Gaussian variational Bayes on the precision matrix}\label{sec:MGVBP}

\subsection{Caveats with the MGVB update}\label{subsec:caveats}
We identify three criticalities concerning the MGVB update as formulated and implemented by \citep{tran_variational_2021}, providing the motivation for this paper and the basis for developing our update.
\begin{itemize}
    \item[(i)] \citep{tran_variational_2021} adopts the manifold of symmetric and positive definite matrices equipped with the Fisher-Rao metric. We denote such manifold by $\mathcal{F}$. 
    As of Section \ref{sec:elements_of_manifold}, in manifold optimization, the update is carried out by employing the retraction function on the Riemann gradient, mapping elements of the tangent space onto the manifold. Indeed, the update \citep{tran_variational_2021} suggests on the variational covariance matrix $\S$ for optimizing the lower bound objective $\LB$, involves
    $R_{\S}\br{\beta \text{grad} \mathcal{\LB}}$.
    Interestingly, they show that for the manifold $\mathcal{F}$, the Riemann Gradient corresponds to the natural gradient (see their Lemma 2.1). However, the form of the retraction they adopt is that of the manifold $\mathcal{M}$ of the SPD matrices equipped with the Euclidean metric and not the retraction form for the manifold $\S$ induced by the Fisher-Rao metric. What is the retraction form for the manifold $\mathcal{S}$ is an open question. 
    In practice, \citep{tran_variational_2021} mixes elements of the two manifolds as it applies the retraction derived from $\mathcal{M}$ on a Riemann gradient of the manifold $\mathcal{F}$.
    This very same point is also raised in \citep{lin2020handling}, which, in fact, underlines that in \citep{tran_variational_2021}, the chosen form of the retraction is not well-justified as it is specific for the SPD matrix manifold, whereas the natural gradient is computed within a different manifold.   
    \end{itemize}
    \citep{tran_variational_2021} establish the equivalence between the Riemann gradient and the natural gradient in $\mathcal{F}$. However, in equations (5.2) and (5.4) of \citep{tan2021natural}, and the implementation accompanying the paper, the natural gradient reads $\S \nabla_\S \mathcal{L} \S$ in place of $2\S \nabla_\S \mathcal{L} \S$ (see Appendix \ref{app:gaussian_fim} and the earlier footnote). This leads to the following two observations:
    \begin{itemize}
    \item[(ii)] The halved natural gradient $\S \nabla_\S \mathcal{L} \S$ is a Riemann gradient for the manifold $\mathcal{M}$, see eq.\eqref{eq:riemann_grad_M_main}. Therefore, the form of retraction \eqref{eq:retraction_general} is coherent and applicable; thus, the update \eqref{eq:tran_updates} on $\S \nabla_\S \mathcal{L} \S$ is formally correct. However, this is a fully consistent procedure for the manifold $\mathcal{M}$, and not for the manifold $\mathcal{F}$ adopted in \citep{tran_variational_2021}.
    \item[(iii)] By adopting $\S \nabla_\S \mathcal{L} \S$ in place of $2\S \nabla_\S \mathcal{L} \S$, in the above light, we recognize that the overall update in \citep{tran_variational_2021}, reads as a hybrid update: whereas $\bmu$ is updated with a natural ingredient update, $\S$ is updated with the gradient $\S \nabla_\S \mathcal{L} \S$, Riemannian in $\mathcal{F}$, but not natural.
\end{itemize}

Regarding point (iii), as for \eqref{eq:analogy}
$\S \nabla_\S \LB \S = \frac{1}{2} \natgrad_\S \LB$. Thus, the Riemann gradient
$\text{grad} \LB$ shares the same direction in both $\mathcal{F}$ and $\mathcal{M}$. Therefore, in both manifolds, the respective Riemann gradients point in the direction of the natural gradient. So the update is, in practice, a natural gradient update. Equation \eqref{eq:analogy}, which does not appear in \citep{tran_variational_2021} and in the earlier VI literature, therefore explains why MGVB actually works despite the above issues.

\begin{figure}
    \centering
    \includegraphics[trim={2.4cm 0.cm 1.4cm 0cm},clip,width = 0.8\textwidth]{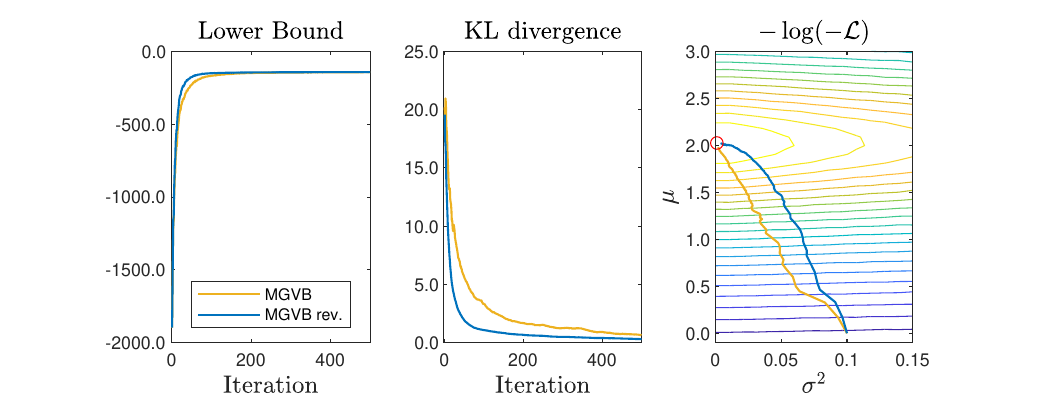}
    \caption{Variational inference for the simple linear regression model, with 100 observations simulated according to $Y = 2X + \bm{\varepsilon}$, $\varepsilon_i \sim \mathcal{N}\br{0,1}$, 
    $X = [0,0.05,0.1,\dots, 5]$, for the MGVB update and the revisited version (MGVB rev. label) applying retraction on the Riemannian gradient in $\mathcal{M}$. 
    The red circle denotes the true posterior parameter computed with standard results in Bayesian linear regression.}
    \label{fig:SLR_low}
\end{figure}
A remedy for the major points (i) and (ii) requires revising the MGVB update for $\S$. Specifically, 
to adopt retraction for $\mathcal{M}$ applied to an actual Riemann gradient for $\mathcal{M}$, i.e., $R_\S(\frac{\beta}{2} \natgrad_\S \LB)$.
Figure \ref{fig:SLR_low} depicts the difference between the MGVB update and such a revisited version. 
The three panels show a simple example of applying retraction \eqref{eq:retraction_general} on the actual Riemann gradient on $\mathcal{M}$, highlighting a non-irrelevant impact on the steps of line-search and the convergence of the algorithm towards the optimum.

The computation of the Riemann gradient $\frac{1}{2}\natgrad_\S \LB = \S \nabla_\S \LB \S$, which involves two matrix multiplications of $\mathcal{O}\br{d^3}$ operations, can, however, be simplified considering a parametrization of the variational distribution in terms of its covariance matrix $\iS$. This leads to our suggested update in the next section. 

\subsection{Updating the precision matrix}
Consider a variational Gaussian distribution with mean $\bmu$ and positive-definite covariance matrix\footnote{\normalsize \doublespacing This is the case of practical relevance, ruling out singular Gaussian distributions for which the discussion is here out of scope. Positive-definiteness is not restrictive and aligned with \citep{tran_variational_2021}.}
$\S$. The corresponding precision matrix $\iS$ is thus well-identified, and the following proposition establishes a central relationship in this regard. 

\begin{proposition}\label{prop:natgrad}
For a $d$-dimensional Gaussian variational posterior whose mean is denoted by $\bmu$ and covariance matrix by $\S$, consider the following two parameterizations: the canonical parameterization $\bz = \br{\bmu^\top,\vec{\S}^\top}^\top$ and the inverse parametrization $\bzp = \br{\bmu^\top,\vec{\iS}^\top}^\top$.
It holds that,
\begin{align}
\natgrad_{\bmu} \LBzp &= \S \nabla_\bmu \LBz\\
\natgrad_{\iS} \LBzp &= -2\nabla_\S \LBz\text{.} \label{eq:5}
\end{align}

\end{proposition}

Equation \eqref{eq:5} establishes an algebraic link between the \textit{natural} gradient of the lower bound in the parametrization $\bzp$ with respect to the precision matrix $\iS$ \--- \enquote{inverse} (covariance) parametrization \--- and the \textit{Euclidean} gradient of the lower bound with respect to respect to the covariance matrix $\S$. In particular, Proposition \ref{prop:natgrad} suggests that under the parametrization $\bzp$, one could update $\mu$ and $\iS$, for which natural gradients are of closed-form solution and correspond to the standard Euclidean gradients of $\LBz$ under the canonical parametrization. 
Compared to the canonical parametrization adopted by \citep{tran_variational_2021} and the discussion in Section \ref{subsec:caveats}, the advantage of using the parametrization $\bzp$ is clear: $\natgrad_{\iS} \LBzp$ requires only the computation of $\nabla_\S \LBz$, whereas $\natgrad_{\iS} \LBzp = \S \nabla_\S \LB \S$ additionally involves additional matrix multiplications, that significantly increase the time-complexity of the lower bound optimization. Section \ref{sec:MGVB:implementation} shows furthermore, such Euclidean gradients are straightforward to compute and that their variance can be conveniently controlled.

The matrix $\iS$ is symmetric and positive definite, thus lies in the manifold $\mathcal{M}$, and can be effectively updated with the retraction algorithm for $\S$ in \eqref{eq:tran_updates}:
\begin{align}\label{eq:update_iS_martin}
   \iS \leftarrow  R_{\iS}\br{\beta \iS \nabla_{\iS} \LBzp \iS} \nonumber &= R_{\iS}\br{\frac{\beta}{2}\natgrad_{\iS}\LBzp}\\ &= R_{\iS}\br{-\beta\nabla_{\S}\LBz}\text{.}
\end{align}
As opposed to \eqref{eq:tran_updates}, updating $\S$ upon the approximation $\I\br{\S} \approx \iS \otimes \iS$ for tacking the positive-definite constraint, we update $\iS$, for which the natural gradient is available in an exact form, avoids the computation of the FIM, and reduces to the simple and standard computation of the Euclidean gradient $\nabla_{\S}\LBz$. 
For updating $\bmu$, it is reasonable to adopt a plain SGD-like step driven by the natural parameter $\natgrad_\bmu \LBzp= \S \nabla_\bmu \LBz$, as in \citep{tran_variational_2021}. 
We refer to the following update rules as Manifold Gaussian Variational Bayes on the Precision matrix (MGVBP):
\begin{align}
    \bmu_{t+1} &= \bmu_t +  \beta \S \nabla_\bmu \LB_t\br{\bz} \text{,} \label{eq:updates_MGVBP_mu}\\
    \iS_{t+1} &= R_{\iS_t}\br{-\beta\nabla_{\S}\LB_t\br{\bz}}  \text{,} \label{eq:updates_MGVBP_S} 
\end{align}
where the gradients are evaluated at the current value of the parameters, e.g., $\nabla_\bmu \mathcal{L}_t\br{\bz} =\nabla_\bmu \LBz\vert_{\bz = \bz_t} $.
Opposed to the MGVB update of \citep{tran_variational_2021}, the update for $\iS$ is consistent for the manifold $\mathcal{M}$, as it employs the corresponding Riemann gradient, $\iS\nabla_{\iS}\LBzp \iS = \nabla_\S \LBz$,  
which is $\mathcal{O}\br{d^3}$ cheaper than computing $\natgrad_\S \LBz = \S \nabla_\S\LBz \S$. 

\subsection{Gradient computations} \label{sec:MGVB:implementation}

We elaborate on how to evaluate the gradients $\nabla_\S\LBz$ and $\nabla_\bmu \LBz$.
We follow the Black-box approach \citep{ranganath2014black} under which such gradients are approximated via Monte Carlo (MC) sampling and rely on function queries only. Implementing MGVBP does not require the model's gradient to be specified nor to be computed numerically, e.g., with backpropagation. 
The so-called log-derivative trick (see, e.g., \citep{ranganath2014black}) makes it possible to evaluate the gradients of the LB as an expectation with respect to the variational distribution. In particular,
\begin{equation*}
    \nabla_\bz\LBz= \E_{q_\bz}\sbr{\nabla_\bz \sbr{\log \qz\brt}\, h_\bz\br\bth}\text{,}
\end{equation*}
where $h_\bz\br{\bth} = \log\sbr{ p\br{\bth} \lik / q_{\bz}\br{\bth}}$.
The gradient can be easily estimated using $S$ samples from the posterior through the unbiased estimator 
\begin{align}\label{eq:naive_mc_grad_estimator}
  \nabla_\bz \LBz &\approx  \frac{1}{S}\sum_{s=1}^S  \sbr{\nabla_\bz\sbr{ \log q_{\bz}\br{\bth_s}} \, h_\bz\br{\bth_s}} \text{,}
\end{align}
with $\bth_s\sim q_{\bz}$.
For the Gaussian variational case under consideration, it can be shown that \citep[e.g.,][]{wierstra2014natural,magris2022quasi}:
\begin{align}
\nabla_\bmu \log q_\bz \br{\bth} &= \iS\br{\bth-\bmu} =\bnu\label{eq:MC_mu} \text{,}\\
\nabla_\S \log q_\bz\br{\bth} &= -\frac{1}{2}\br{\iS - \bnu \bnu^\top} \label{eq:MC_S} \text{.}
\end{align}
Equations \eqref{eq:MC_mu}, \eqref{eq:MC_S} along with \eqref{eq:naive_mc_grad_estimator} and Proposition \ref{prop:natgrad} immediately lead to the feasible natural gradients estimators:
\begin{align*}
  \natgrad_{\bmu}\LBzp  &\approx \S_t\hat{\nabla}_\bmu \LBz 
  = \frac{1}{S}\sum_{s=1}^S \sbr{ \br{\bth_s-\bmu} h_{\bz}\br{\bth_s} } \text{,} \\
    \natgrad_{\iS} \LBzp &\approx -\hat{\nabla}_\S \LBz   \nonumber =\frac{1}{2S} \sum_{s=1}^S  \sbr{\br{\iS-\bnu_{s}{\bnu_{s}}^\top} h_{\bz}\br{\bth_s}}\text{,}
\end{align*}
with $\bnu_{s} = \iS\br{\bth_s-\bmu}$.
As for the MGVB update, MGVBP applies exclusively to Gaussian variational posteriors, yet no constraints are imposed on the parametric form of the prior $p\brt$. When considering a Gaussian prior, the implementation of the MGVBP update can take advantage of some analytical results leading to MC estimators of reduced variance, namely implemented over the log-likelihood $\log p\br{\by\vert \bth_s}$ rather than the $h$-function.

In Appendix \ref{app:proof_MGVBP_cases}, we show that under a Gaussian prior specification, the above updates can also be implemented in terms of the model likelihood rather than the $h$-function.  
At iteration $t$, the general form of the gradients evaluated at the current (epoch-$t$) values of the parameters read
\begin{align}
      \natgrad_{\bmu} \LB_t\br{\bzp} &\approx  c_{\bmu_t} + \frac{1}{S}\sum_{s=1}^S \sbr{ \br{\bth_s-\bmu_t} \log f\br{\bth_s} } \text{,} \label{eq:natgrad_mu_llh}\\
      \natgrad_{\iS} \LB_t\br{\bzp}  &\approx  C_{\S_t} + \frac{1}{2S} \sum_{s=1}^S\sbr{\br{\iS_t-\bnu_{t,s}{\bnu_{t,s}}^\top} \log f\br{\bth_s} }\text{,} \label{eq:natgrad_S_llh} 
\end{align}
with $\bnu_{s,t} = \iS_t\br{\bth_s-\bmu_t}$ where in general (whether the prior is Gaussian or not)

\begin{equation}\label{eq:estim_general}
    \begin{cases}
        C_{\S_t} = 0 \text{,}\\
        c_{\bmu_t} = \bm{0} \text{,}\\  
        \log f\br{\bth_s} = h_{\bm{\zeta}^c_t}\br{\bth_s} \text{,}
    \end{cases}
\end{equation}
whereas for a Gaussian prior, one can adopt
\begin{equation}\label{eq:estim_gaussian}
    \begin{cases}
        C_{\S_t} = -\frac{1}{2}\iS_t +\frac{1}{2}\iSz \text{,}\\
        c_{\bmu_t} = -\S_t \iSz\br{\bmu_t-\bmu_0} \text{,}\\
        \log f\br{\bth_s} = \log p\br{\by\vert \bth_s} \text{.}
    \end{cases}
    \end{equation}

That is, \eqref{eq:estim_general} holds for a generic prior (including a Gaussian prior as a special case), whereas \eqref{eq:estim_gaussian} holds with Gaussian priors only.
As shown in Appendix \ref{app:proof_MGVBP_cases}, under a Gaussian prior, certain components of the $h$-function involved in \eqref{eq:estim_general} have an algebraic solution, and the MC estimator \eqref{eq:estim_gaussian} based on the log-likelihood function is of reduced variance. Thus, under a Gaussian prior, \eqref{eq:estim_gaussian} is preferred to \eqref{eq:estim_general}. This alternative estimator is applicable in \citep{tran_variational_2021} and, generally, in other black-box Gaussian VI contexts.
Note that the log-likelihood case does not involve an additional inversion for retrieving $\S$ in $c_\bmu$, as $\S$ is anyway required in the second-order retraction (for both MGVB and MGVBP). This aspect is further developed in Section \ref{subsec:computational}. For inverting $\iS$ we suggest inverting the Cholesky factor $L$ of $\iS$ and compute $\S$ as $L^{-\top} L^{-1}$. The triangular form of $L$ can be inverted with back-substitution, requiring $d^3/3$ flops instead of $d^3$. $L^{-\top}$ is furthermore used for generating the draws $\bth_s$ as $\bth_s = \bmu + L^{-\top}\bm{\varepsilon}$, with $\bm{\varepsilon} \sim \N\br{\bm{0},I}$. We suggest using control variates to reduce the variance of the stochastic gradient estimators; see Section \ref{subsec:cv}.

Though the lower bound is not directly involved in MGVBP updates, it can be naively estimated at each iteration as 
\begin{equation}\label{eq:MC_LB}
    \hat{\LB}_t\br{\bz} = \frac{1}{S}\sum_{s=1}^S\sbr{\log p\br{\bth_s} + \log p\br{\by \vert \bth_s} - \log q_{\bm {\zeta}^c_t} \br{\bth_s}}\text{,} \quad \bth \sim q_{\bm{\zeta}^c_t}\text{.}
\end{equation}
$\hat{\LB}_t\br{\bz}$ is required for terminating the optimization routine (see Section \ref{subsec:stopping_crit}), verifying anomalies in the algorithm (the LB should increase across the iterations and eventually and converge), and comparing MGVBP with MGVB, as in Section \ref{sec:experiments}.

\subsection{Retraction and vector transport} \label{subsec:MGVBP_retr_and_vec_transport}
Aligned with \citep{tran_variational_2021}, we adopt the retraction method advanced in \citep{jeuris2012survey} for the manifold $\mathcal{M}$, but on the actual Riemannian gradients for this manifold,
\begin{equation}\label{eq:retr}
    R_{\iS}\br{\bxi} = \iS +\bxi +\frac{1}{2} \bxi \S \bxi \text{,}
\end{equation}
with $\bxi \in T_{\iS} \mathcal{M}$ being the rescaled natural gradient $\beta/2\natgrad_{\iS}\LBz = -\beta \nabla_\S \LBz$.
Vector transport is easily implemented by
\begin{equation}\label{eq:transport}
 \mathcal{T}_{\iS_t \rightarrow \iS_{t+1}}\br{\bxi} = E \bxi E^\top \text{,} 
\end{equation}
with $E = \br{\iS_{t+1} \S_{t}}^\frac{1}{2}, \, \bxi \in T_{\iS} \mathcal{M}$.
In implementations, for numerically granting the symmetric form of a matrix $P$, we compute $P$ as $1/2(P+P^\top)$. We refer to the Manopt toolbox \citep{manopt2014} for further practical details on implementing the above two algorithms in a numerically stable fashion. The momentum gradients immediately follow:
\begin{align}
\natgrad_{\iS}^{\text{mom.}}\LB_{t+1}\br{\bzp} =  \omega\,& \mathcal{T}_{\iS_t \rightarrow \iS_{t+1}}\br{\natgrad_{\iS}^{\text{mom.}}\LB_{t}\br{\bzp}} +\br{1-\omega} \natgrad_{\iS}\LB_{t+1}\br{\bzp}\text{,} \label{eq:mom_mu}\\
\natgrad_{\bmu}^{\text{mom.}}\LB_{t+1}\br{\bzp} = \omega\,& \natgrad_{\bmu}^{\text{mom.}}\LB_t\br{\bzp} + \br{1-\omega}\natgrad_\bmu \LB_t\br{\bzp} \text{,}   \label{eq:mom_iS}
\end{align}
where the weight $0<\omega<1$ is a hyper-parameter.
Algorithm \ref{alg:MGVBP_hfunc} summarizes the MGVBP update for the Gaussian prior-variational posterior case. Aspects of relevance in its implementation are discussed in Section \ref{sec:implementation_aspects}.

\begin{center}
\begin{minipage}{0.8\linewidth}

\begin{algorithm}[H]
\centering
\caption{MGVBP implementation}\label{alg:MGVBP_hfunc}
\begin{algorithmic}
   \STATE \text{Set hyper-parameters:\,}$ 0<\beta,\omega< 1$, $S$
   \STATE \text{Set the type of gradient estimator, i.e. function $\log f\br{\bth_s}$}
\STATE \text{Set the initial values for $\bmu$, $\iS$, specify $p\brt$ and $p\br{\by\vert\bth}$}
\STATE $t=1$, $\text{Stop} = \texttt{false}$
\STATE \text{Generate:\,} $\bth_s \sim q_{\bmu,\S}$, $s = 1,\dots, S$
\STATE \text{Compute:\,} $\hat{g}_\bmu = \S\hat{\nabla}_\bmu \LB$, $\hat{g}_{\iS} = -\hat{\nabla}_{\S}\LB $ \label{eq:alg_mu} \COMMENT{eqs.\,\eqref{eq:natgrad_mu_llh},\,\eqref{eq:natgrad_S_llh}}
\STATE $m_\bmu = \hat{g}_\bmu$, $m_{\iS} = \hat{g}_{\iS}$ \COMMENT{initialize momentum}
\WHILE{$\text{Stop} = \texttt{false}$}
\STATE $\bmu = \bmu + \beta m_\bmu $ \COMMENT{MGVBP update for $\bmu$}
\STATE $\iS_{\text{old}} = \iS$
\STATE $\iS = R_{\iS_{\text{old}}}\br{\beta m_{\iS}}$ \COMMENT{MGVBP update for $\iS$}
\STATE \text{Generate:\,} $\bth_s \sim q_{\bmu,\S}$, $s = 1,\dots, S$ 
\STATE \text{Compute:\,} $\hat{g}_\bmu$, $\hat{g}_{\iS}$ 
\STATE $m_{\bmu} = \omega m_\bmu + \br{1-\omega} \hat{g}_\bmu$ \COMMENT{eq.\eqref{eq:mom_mu}}
\STATE $m_{\iS} = \mathcal{T}_{\iS_{\text{old}} \rightarrow \iS}\br{m_{\iS}}+\br{1-\omega} \hat{g}_{\iS}$ \COMMENT{eq.\,\eqref{eq:mom_iS}}
\STATE \text{Compute:\,} $\hat{\LB}$ \COMMENT{eq.\,\eqref{eq:MC_LB}}
\STATE $t = t+1$, $\text{Stop} =  f_{\text{exit}}(t,\dots)$ \COMMENT{see Section \ref{sec:implementation_aspects}}
\ENDWHILE
\end{algorithmic}
\end{algorithm}

\end{minipage}
\end{center}

\subsection{Isotropic prior}

For mid-sized to large-scale problems, the prior is commonly specified as an isotropic Gaussian of mean $\bmu_0$, often $\bmu_0 = \bm{0}$, and covariance matrix $\tau^{-1} I_d$, with $\tau>0$ a scalar precision parameter.
The covariance matrix of the variational posterior can be either diagonal or not. Whether a full covariance specification ($d^2$ parameters) can provide additional degrees of freedom to gauge models' predictive ability, a diagonal posterior ($d$ parameters) can be practically and computationally convenient to adopt, e.g., in large-sized problems. The diagonal-posterior assumption is broadly adopted in Bayesian inference and VI, e.g., \citep{blundell2015weight,ganguly2021introduction,tran2021practical} and Bayesian ML applications, e.g., \citep{kingma2013auto,graves2011practical,khan2018fastscalable,osawa_practical_2019}, in Appendix \ref{app:further_cons} we provide a block-diagonal variant.

\subsubsection*{Isotropic prior and diagonal Gaussian posterior}
Assume a $d$-variate diagonal Gaussian variational specification, that is $q \sim \N\br{\bmu,\S}$ with $\diag{\S} = \s$, $\S_{ij}=0$, for $i,j = 1,\dots,d$ and $i\neq j$. In this case, $\iS = \diag{1/\s}$, where the division is intended element-wise, is now a $d \times 1$ vector. 
Be $\bz = \br{\bmu^\top,\br{\s}^\top}^\top$
and $\bzp = \br{\bmu^\top,\br{\is}^\top}^\top$,
thus $\nabla_{\s} \LBz =  \diag{\nabla_{\s} \LBz}$. Updating $\iS$ amounts to updating $\is$: the natural gradient retraction-based update for $\is$ is now based on the equality $\natgrad_{\is} \LBzp = -\nabla_{\s} \LBz$, so that the general-case MGVBP update reads
\begin{align*}
\bmu_{t+1} &= \bmu_t + \s_t \odot \beta \nabla_\bmu \LBz \textrm{,}\\
\is_{t+1} &= R_{\bm{\sigma}^{-2}_{t+1}}\br{- \beta \nabla_{\s} \LBz} \textrm{,} 
\end{align*}
where $\odot$ denotes the element-wise product, and the retraction is adapted to   
$$R_{\bm{\sigma}^{-2}_t}\br{\bxi} = \bm{\sigma}^{-2}_t + \bxi +\frac{1}{2}\bxi \odot \bm{\sigma}^{-2}_t \odot\bxi \text{,} $$
where $\bxi$ is a $d-$dimensional vector.
The corresponding MC estimators for the gradients are
\begin{align*}
  \natgrad_\bmu \LB_t \br{\bzp}\, &\approx \s \odot \hat{\nabla}_\bmu \LB_t\br{\bz}\\
  &=  c_{\bmu_t} + \frac{1}{S}\sum_{s=1}^S \sbr{\br{\bth_s-\bmu_t} \log p\br{\by\vert \bth_s} } \text{,}\\
  \natgrad_{\is} \LB_t \br{\bzp}\, &\approx -\hat{\nabla}_{\s} \LB_t\br{\bz}\\
  &= c_{\s_t} +\is_t \odot \frac{1}{2S} \sum_{s=1}^S \sbr{ \br{\bm{1}_d - \br{\bth_s - \bmu_t}^2  \odot \is_t}  \log p\br{\by\vert \bth_s} }  \text{,}  \nonumber
\end{align*}
where $c_{\s_t} = - 1/2\is_t + \tau/2$, $c_{\bmu_t} = \tau \s_t \odot \br{\bmu_t-\bmu_0} $, $\bth_s \sim \N\br{\bmu_t,\diag{\s_t}}$, $s = 1\dots,S$, $\br{\bth_s - \bmu_t}^2$ is intended element-wise, and $\bm{1}_d = \br{1,\dots,1}^\top \in \R^d$. 
In the Gaussian case with a general diagonal covariance matrix, retrieving $\s$ from the updated $\is$ is inexpensive as $\sigma^{2}_{i} = 1/\sigma^{-2}_{i}$.

\subsubsection*{Isotropic prior and full Gaussian posterior}

Because of the full form of the covariance matrix, this case is rather analogous to the general one. In particular, factors  $c_{\bmu_t}$ and $c_{\S_t}$ in eq. \eqref{eq:cases} are replaced by
(i) $c_{\S_t} = -1/2\iS_t +\tau/2$, $c_{\bmu_t} = \tau \S_t \br{\bmu_t-\bmu_0}$ or
(ii) $c_{\S_t} = 0$, $c_{\bmu_t} = 0$, respectively, under the Gaussian-prior case ($\log f\br{\bth_s} = \log p\br{\by\vert \bth_s}$) and the general one ($\log f\br{\bth_s} = h \br{\bth}$). The MC estimators \eqref{eq:MC_mu} and \eqref{eq:MC_S} apply but are of reduced variance.

\subsection{Mean-field variant} \label{subsec:mean_field}
Assume that for a $d$-variate model, the Gaussian variational posterior can be factorized as
$$
     q_{\bz}\br{\bth} = q_{\bm{\zeta}^c_1}\br{\bth_1}q_{\bm{\zeta}^c_2}, \dots, q_{\bm{\zeta}^c_h}\br{\bth_h} = \prod_{j=1}^h q_{\bm{\zeta}^c_j}\br{\bth_j}\text{,}
$$ with $h\leq d$.
If $h=d$, this corresponds to a full-diagonal case where each $\bth_j$ is a scalar and $\bz_j = \br{\mu_j,\sigma_j}$. 
If $h<d$, the variational covariance matrix $\S$ of $q_{\bz}$ corresponds to a block-diagonal matrix, and $\bm{\zeta}^c_j = \br{ \bmu_j^\top,\vec{\S_j}^\top}^\top$, with $\bmu_j$, $\S_j$ ($\iS_j$) respectively denoting the mean and covariance (precision) matrix of the $i$-th block of $\S$ ($\iS$).
Also, a Gaussian prior's covariance matrix can be diagonal, full, or block-diagonal with a structure matching or not that of $\S$. Equations \eqref{eq:natgrad_mu_llh}, \eqref{eq:natgrad_S_llh}, with the condition \eqref{eq:cases} can be used as a starting point to derive case-specific MGVBP variants based on the form of the prior covariance.

Algorithm \ref{alg:mean_field} summarizes the case with an isotropic Gaussian prior of zero-mean and precision matrix $\tau I_d$, using the gradient estimator based on the log-likelihood. In this case, the block-wise natural gradients are estimated as 
\begin{align*}
\S\hat{\nabla}_{\bmu_j}\LB\br{\bm{\zeta}^c_j} &=  -\tau\iS_j\bmu_j + \frac{1}{S}\sum_{s=1}^S \sbr{ \br{\bth_{s_j}-\bmu_j}
 \log p\br{\by\vert \bth_{s_j}} } \text{,}\\
-\hat{\nabla}_{\S_j}\LB\br{\bm{\zeta}^c_j}  &= -\frac{1}{2}\iS_j+ \frac{\tau}{2} I + \frac{1}{2S} \sum_{s=1}^S\sbr{\br{\iS_j-\bnu_j \bnu_j^\top} \log p\br{\by\vert \bth_{s_j}} }\text{,}
\end{align*}
where $I$ is the identity matrix of appropriate size for the block $j$, $\bnu_{j} = \iS_j\br{\bth_{s_j}-\bmu_j}$, and $\bth_{s_j} \sim q_{\bm{\zeta}^c_j}$, $j = 1,\dots,h$.

\subsection{Computational aspects}\label{subsec:computational}
In terms of computational complexity, the exact MGVBP implementation is at no additional cost. Actually, the cost of computing the natural gradient in MGVBP as $-\nabla_{\S}\LBz$ is much cheaper than the one in MGVB, $\S\nabla_\S\LBz\S$, involving $O\br{d^3}$ operations for each matrix multiplication. However, both MGVB and MGVBP share a cumbersome matrix inversion.

Going back to \eqref{eq:natgrad_S_llh} and \eqref{eq:natgrad_mu_llh}, it is noticeable that under the most general estimator based on the $h$-function, $\S$ is not involved in any computation, neither in the gradient involved in the update for $\iS$ nor in that for $\bmu$, suggesting that implicitly the MGVBP optimization routine does not require the inversion of $\iS$. 
The above point, however, ignores that the underlying retraction masks the update for $\iS$. For the retraction form in eq. \eqref{eq:retr}, both $\iS$ and its inverse $\S$ are needed, thus implying a matrix inversion at every iteration. That is, the covariance matrix inversion is implicit in MGVB and MGVBP methods, which both require $\iS$ and $\S$ at every iteration (with little surprise, as the form of the retraction is a second-order approximation of the exponential map). 

With the $h$-function estimator, even though neither \eqref{eq:MC_mu} nor \eqref{eq:MC_S} involve $\S$, the inversion of $\iS$ is still necessary, as $\S$ is required in retraction. Similarly, adopting the log-likelihood estimator under the Gaussian regime in \eqref{eq:cases}, is not computationally more expensive than the $h$-function case, as $\S$, involved in the computation of $c_{\bmu}$, is, once again, required by retraction. 
In Appendix \ref{app:exp:hyperp}, we compare the running times of MGVBP and MGVB, showing that they are indeed rather aligned, despite the computational simplicity of the MGVBP natural gradient.

As outlined in Section \ref{sec:MGVBP}, $\S$ can be conveniently recovered from the Cholesky factor of $\iS$, with fewer flips. Lastly, if $\iS$ (or $\S$) is diagonal, the inversion is trivial and, when applicable, \eqref{eq:natgrad_mu_llh} is preferred. 

\section{Implementation aspects}\label{sec:implementation_aspects}

\subsection{Classification vs. regression}
We point out that the MGVBP framework applies to both regression and classification problems. In generic classification problems, predictions are based on the class of maximum probability, obtained through a softmax function returning the $K \times 1$ probability vector $\bm{p}_i$, whose $k$-th entry interprets as the probability that the label of the $i$-th sample is in class $k$, $k\in\{1,\dots,K\}$. From these probabilities, it is straightforward to compute the model log-likelihood as $\sum_{i=1}^N \by_{i} \log \bm{p}_i$, where $\by_{i}$ denotes the one-hot-key encoded true label for the $i$-th input: a $1 \times K$ vector with 1 at the element corresponding to the true label and 0 elsewhere. 

For regression, the parametric form of $\log p\br{\by \vert \bth}$ is clearly different and model-specific (e.g., regression with normal errors as opposed to Poisson regression, with the latter being feasible as the use of the score estimator does not require the likelihood to be differentiable). Note that, however, additional parameters may enter into play besides the ones involved in the back-bone forward model. For instance, for a regression with normal errors tackled with an artificial neural network, the Gaussian likelihood involves the regression variance, which is an additional parameter over the network's ones, or the degree of freedom parameter $\nu$ (with the constraint $\nu>2$) for Student-t errors. See the application in Appendix \ref{app:istanbul}.

\subsection{Variance reduction}\label{subsec:cv}
As MGVBP does not involve model gradients, the use of the Reparametrization Trick (RT) \citep{blundell2015weight} is not immediate. While \eqref{eq:updates_MGVBP_mu} and \eqref{eq:updates_MGVBP_S} would generally hold, the form of the MGVBP gradient estimators under the RT would differ from \eqref{eq:MC_mu} and \eqref{eq:MC_S}. We develop MGVBP as a general and ready-to-use solution for VI that does not require model-specific derivations, yet one may certainly enable the RT within MGVBP. The use of the RT is quite popular in VI and ML as it empirically yields more accurate estimates of the gradient of the variational objective than alternative approaches \citep{mohamed2020monte}. However, note that, in general, the use of the RT estimator is not necessarily preferable, as its variance can be higher than that of the score-function estimator \citep{xu2019variance,mohamed2020monte}. Furthermore, the applicability of the adopted score estimator is broader, as it does not require $\log p\br{\by \vert \bth}$ to be differentiable.

Control Variates (CV) is a simple and effective approach for reducing the variance of the MC gradient estimator. The CV estimator
$$
\frac{1}{S}\sum_{s=1}^{S} \nabla_{\bz}\sbr{\log q_\bz\br{\bth_s}}  \br{\log p\br{\by\vert \bth_s} - c} \text{,}
$$
is unbiased for the expected gradient but of equal or smaller variance than the naive MC one. The optimal $c$ minimizing the variance of the CV estimator is \citep[see, e.g.,][]{paisley2012variational,ranganath2014black}
\begin{equation}\label{eq:c_star}
    c^\star = \frac{\Cov \br{\nabla_{\bz}\sbr{\log q_\bz\br{\bth}} \log p\br{\by \vert \bth},\nabla_{\bz}\log q_\bz\br{\bth}}}{\Var\br{\nabla_{\bz}\log q_\bz\br{\bth}}} \text{.}
\end{equation}
For a given $S$, the CV estimator reduces the variance of the gradient estimates or, conversely, reduces the number of MC samples required for attaining a desired level of variance. In Table \ref{tab:logistic_S}, we asses that for logistic regression, values of $S$ as little as 10 appear satisfactory. Yet, if the iterative computation of the log-likelihood is not prohibitive, we suggest adopting a more generous value. \citep{magris2022quasi} furthermore shows that the denominator in \eqref{eq:c_star} is analytically tractable for a Gaussian choice for $q$, reducing the variance of the estimated $c^\star$ and thus improving the overall efficiency of the CV estimator.

\subsection{Constraints on model parameters}
MGVBP assumes a Gaussian variational posterior: the mean parameter is unbounded and defined over the entire real line. Assuming that a model parameter $\bth$ is required to lie on a support $\mathcal{S}$, to impose such a constraint, it suffices to identify a feasible transform $T: \R \rightarrow \mathcal{S} $ and apply the MGVBP update to the unconstrained parameter $\bm{\psi} = T^{-1}\br{\bth}$. By applying VI on $\bm{\psi}$, we require that the variational posterior assumption holds for $\bm{\psi}$ rather than $\bth$. The actual distribution for $\bth$ under a Gaussian variational distribution for $\bm{\psi}$ can be computed (or approximated with a sampling method) as $\N\br{T^{-1}\br{\bth};\bmu,\S}\vert \det \br{J_{T^{-1}}\br{\bth}}\vert$, with $J_{T^{-1}}$ the Jacobian of the inverse transform, and $\N$ denoting the multivariate normal probability density function \citep{kucukelbir2015automatic}.

\noindent
\textit{Example.} For the GARCH(1,1) model, the intercept $\omega$, the autoregressive coefficient of the lag-one squared return $\alpha$, and the moving-average coefficient $\beta$ of the lag-one conditional variances need to satisfy the stationarity condition $\alpha +\beta <1$ with $\omega >0$, $\alpha \geq 0$, $\beta \geq0$. Such conditions are unfeasible under a Gaussian variational approximation. As in \citep{magris2023variational}, we estimate the unconstrained parameters $\psi_\omega, \psi_\alpha, \psi_\beta$, where $\omega = T\br{\psi_\omega}, \alpha = T\br{\psi_\alpha}\br{1-T\br{\psi_\beta}}, \beta = T\br{\psi_\alpha}T\br{\psi_\beta}$ with $T\br{x} = \exp\br{x}/\br{1+\exp\br{x}}$ for $x$ real, on which Gaussian prior-posterior assumptions apply.

\subsection{LB smoothing and stopping criterion}\label{subsec:stopping_crit}
The stochastic nature of the lower bound estimator \eqref{eq:MC_LB} introduces some noise that can violate the expected non-decreasing behavior of the lower bound across the iterations. By setting a window of size $w$ we compute the moving average $\bar{\LB}_t = 1/w \sum_{i=1}^w \hat{\LB}_{t-i+1}\br{\bz}$, whose variance is reduced, and behavior stabilized. 

By keeping track of $\LB^\star := \max \bar{\LB}_t$, occurring at some iteration $t^\star$, $1\leq t^\star \leq t$, we terminate the optimization after $\LB^\star$ did not improve for $P$ iterations (patience parameter), or after a maximum number of iterations $t_{\text{max}}$ is reached. 
Therefore, the termination of the algorithm is determined by an exit condition which depends on $t$, $t_{\text{max}}$, the distance $t-t^\star$, the values of $\hat{\LB}_{t-w+1},\dots,\hat{\LB}_t$, and $\LB^\star$. Here $t_{\text{max}},P$ and $w$ are hyperparameters. In algorithms \ref{alg:MGVBP_hfunc} and \ref{alg:mean_field}, this exit function is compactly denoted by $f_{\text{exit}}\br{t,\dots}$.

\subsection{Gradient clipping}
Especially for low values of $S$, and even more, if a variance control method is not adopted, the stochastic gradient estimate may be poor, and the offset from its actual value may be large. This can result in updates whose magnitude is too big, either in a positive or negative direction. Especially at early iterations and with poor initial values, this issue may, e.g., cause complex roots in \eqref{eq:transport}. At each iteration $t$, to control for the magnitude of the stochastic gradient $\hat{\bm{g}}_t$, we rescale its $\mathcal{\ell}_2$-norm $\vert \vert \hat{\bm{g}}_t \vert \vert$  whenever it is larger than a fixed threshold $l_\text{max}$ by replacing $\hat{\bm{g}}_t$ with $\hat{\bm{g}}_t l_\text{max}/\vert\vert \hat{\bm{g}}_t \vert \vert$, which bounds the norm of the rescaled gradient, while preserving its direction. Gradient clipping can be either applied to the gradients $\hat{\nabla}_\bmu \LBz$, $\hat{\nabla}_\S \LBz$ or/and to the natural gradients $\S\hat{\nabla}_\bmu \LBz$, $-\hat{\nabla}_\S \LBz$. We suggest applying gradient clipping on both gradients to promptly mitigate the impact that far-from-the-mean estimates may have on natural gradient computations and to control the norm of the product $\S\hat{\nabla}_\bmu \LBz$.

\subsection{Adaptive learning rate}
Adopting an adaptive learning rate or scheduler for decreasing the learning rate $\beta$ after a certain number of iterations is convenient. Typical choices include multiplying $\beta$ by a certain factor (e.g., 0.2) every set number of iterations (say, 100) or dynamically updating it after a certain iteration $t'$. We adopt $\beta_t = \min(\beta,\beta \frac{t'}{t})$, where $t'$ is a fraction (e.g., 0.7) of the maximum number of iterations $t_{\text{max}}$.

\section{Experiments}\label{sec:experiments}
We validate and explore our suggested optimizer's empirical validity and feasibility over five datasets and 14 models. These include logistic regression (Labor dataset), different volatility models on the S\&P 500 returns (S\&P 500 dataset), linear regression on different stock indexes (Istanbul data), a neural network (Limit-Order Book (LOB) data), and a non-differentiable model (Synthetic data). Details on the datasets and models appear in Table \ref{tab:hyperparameters}.
Our experiments deal with both classification and regression tasks.
For classification, we discuss binary-class prediction with logistic regression and time-series multi-class classification with a neural network. For regression, we propose a linear model and several GARCH-like models for volatility modeling. Lastly, we exploit the advantage of the black-box setting for regression within a non-differentiable model. Extended results, with additional datasets and models, appear in Appendix \ref{app:experiments}. The relevant codes are available at \url{github.com/mmagris/MGVBP}.
The main baseline for model comparison is the MGVB optimizer and (sequential) MCMC estimates representative of the true posterior. Additionally, we also include results related to the QBVI optimizer \citep{magris2022quasi}, Maximum Likelihood (ML), and ADAM when applicable.

\subsection{Classification tasks}

The logistic regression experiments provide a sanity and qualitative check on the feasibility, robustness, and learning process of MGVBP. From the convex form of the likelihood, we expect to observe a tight alignment of the parameters' estimates and performance metrics due to the similar optimum the different algorithms attain for the relatively simple form of the LB under this experimental setting. This is indeed the case for the results presented later. In such a setting, we can grasp and largely comment on the qualitative differences in the learning process and final results obtained with the different optimizers. 

\begin{figure}
    \centering
    \includegraphics[trim={2cm 0cm 1cm 0cm}, clip, width = 1.05\textwidth]{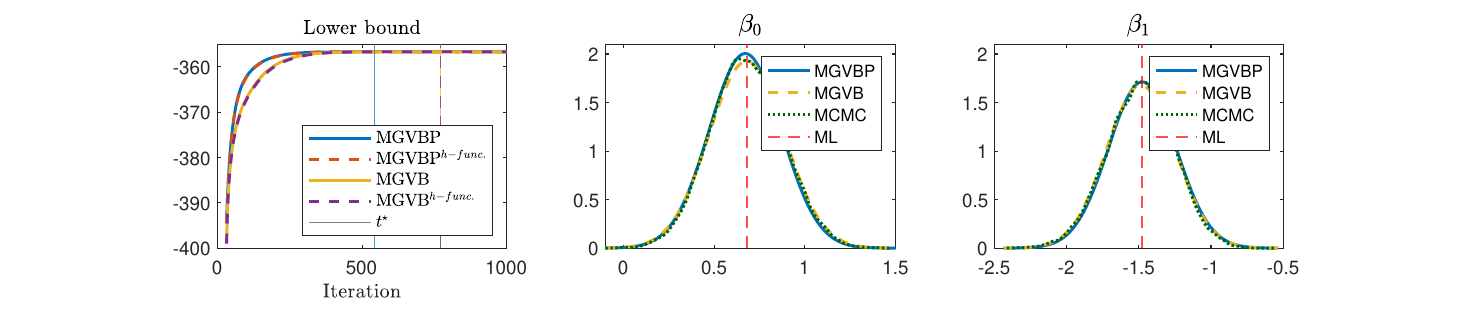}
    \caption{Logistic regression. Left: dynamics of the lower bound across the iterations. Center and right: marginal posteriors for the mode internet and coefficient of the first regressor for MGVBP, MGBV. A kernel density from MCMC samples and the ML solutions are overlaid.}
    \label{fig:labour_LBmargins}
\end{figure}

\begin{figure}
    \centering
    \includegraphics[trim={2cm 0cm 1cm 0cm}, clip, width = 1.05\textwidth]{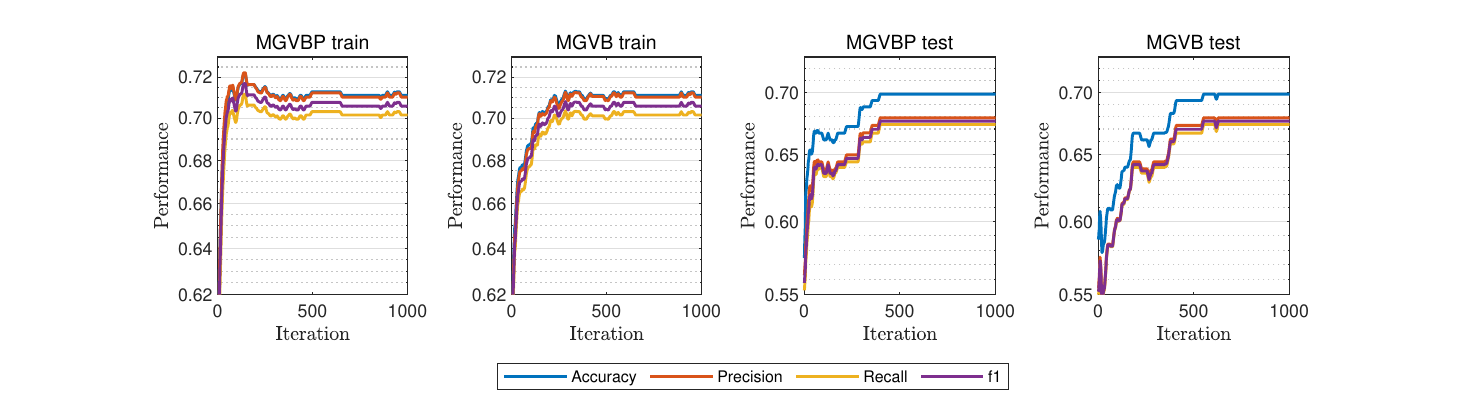}
    \caption{Logistic regression. Dynamics of the performance metrics across the iterations on the train and test set for the MGVBP and MGVB optimizers.}
    \label{fig:Labour_Perf}
\end{figure}

The upper panels in Figure \ref{fig:labour_LBmargins} depict the LB optimization process across the iterations for the logistic regression. MGVBP's LB shows a steeper growth rate and improved convergence rate on the training set, irrespective of whether the $h$-function estimator is used. The central and left panels of Figure \ref{fig:labour_LBmargins} furthermore highlight an excellent accordance of the learned MGVBP variational posterior with its MCMC counterpart, with means aligned with the ML estimates. This underlines that the Gaussian variational assumption is well-suited and that the moments of the variational marginal match those derived from the MCMC sampler. 
In Figure \ref{fig:labour_LBmargins}, we plot the dynamics of standard performance metrics for classification tasks. On both the training and test set, at early iterations, MGVBP displays a very steep growth in model accuracy, precision, recall, and f1-score compared to MGVB. Eventually, at later iterations, MGVBP and MGVB performance metrics converge to a similar level. Yet, as, e.g., depicted by the vertical lines in the left panel of Figure \ref{fig:labour_LBmargins}, MGVBP leads to LB convergence approximately 200 iterations earlier than MGVB, reflected in the values of the performance metrics. That is, the training is completed in a much smaller number of iterations. 

In fact, this is aligned with \ref{tab:logistic_perf}, showing that the LB reaches a similar maximum value corresponding to rather analogous performance metrics. As a consequence, the final variational solution MGVBP attains with respect to the existing optimizers is similar. Table \ref{tab:logistic_mu} reports the estimated posterior means, \ref{tab:logistic_var} the variances, and Table \ref{tab:logistic_cov} the estimated covariances.

\begin{table*}[ht]
    \centering
    \caption{Performance metrics and value of the optimized lower bound ($\LB^\star$) for all the classification experiments on the Labour dataset. $^{h\text{-func.}}$ denotes the use of the $h$-function gradient estimator, and $^\text{diag.}$ the use of a diagonal variational posterior.}
  \scalebox{0.57}{ 
    \begin{tabular}{lSSSSSSSSS}&\\
          &  \multicolumn {5}{c}{Train}             & \multicolumn {4}{c}{Test} \\
\cmidrule(lr){2-6}  \cmidrule(lr){7-10}       & \multicolumn {1}{c}{$\LB^\star$} & \multicolumn {1}{c}{Accuracy} & \multicolumn {1}{c}{Precision} & \multicolumn {1}{c}{Recall} & \multicolumn {1}{c}{f1}    & \multicolumn {1}{c}{Accuracy} & \multicolumn {1}{c}{Precision} & \multicolumn {1}{c}{Recall} & \multicolumn {1}{c}{f1}\\
    \toprule
    MGVBP & -356.645 & 0.713 & 0.712 & 0.703 & 0.708 & 0.698 & 0.679 & 0.674 & 0.676 \\
    MGVB  & -356.646 & 0.711 & 0.710 & 0.701 & 0.706 & 0.698 & 0.679 & 0.674 & 0.676 \\
    QBVI  & -356.645 & 0.713 & 0.712 & 0.703 & 0.708 & 0.698 & 0.679 & 0.674 & 0.676 \\
    MGVBP$^{h\text{-func.}}$ & -356.635 & 0.711 & 0.710 & 0.701 & 0.706 & 0.698 & 0.679 & 0.674 & 0.676 \\
    MGVB$^{h\text{-func.}}$ & -356.635 & 0.711 & 0.710 & 0.701 & 0.706 & 0.698 & 0.679 & 0.674 & 0.676 \\
    QBVI$^{h\text{-func.}}$ & -356.635 & 0.711 & 0.710 & 0.701 & 0.706 & 0.698 & 0.679 & 0.674 & 0.676 \\
    MGVBP$^\text{diag.}$ & -358.605 & 0.713 & 0.712 & 0.703 & 0.708 & 0.672 & 0.650 & 0.647 & 0.649 \\
    MGVB$^\text{diag.}$ & -358.681 & 0.709 & 0.708 & 0.701 & 0.704 & 0.667 & 0.645 & 0.643 & 0.644 \\
    QBVI$^\text{diag.}$ & -358.594 & 0.713 & 0.712 & 0.703 & 0.708 & 0.672 & 0.650 & 0.647 & 0.649 \\    
    MGVBP$^{h\text{-func. diag}}$ & -358.601 & 0.713 & 0.712 & 0.703 & 0.708 & 0.667 & 0.644 & 0.640 & 0.642 \\
    MGVB$^{h\text{-func. diag}}$ & -358.680 & 0.711 & 0.710 & 0.702 & 0.706 & 0.667 & 0.644 & 0.640 & 0.642 \\
    QBVI$^{h\text{-func. diag}}$ & -358.590 & 0.713 & 0.712 & 0.703 & 0.708 & 0.667 & 0.644 & 0.640 & 0.642 \\
    \bottomrule
    \end{tabular}%
    }
\label{tab:logistic_perf}
\end{table*}

\begin{table*}[ht]
    \centering
    \caption{Posterior variational means for all the classification experiments on the Labour dataset.}
\scalebox{0.7}{    
\begin{tabular}{lSSSSSSSS}
      & \multicolumn {1}{c}{$\beta_0$} & \multicolumn {1}{c}{$\beta_1$} & \multicolumn {1}{c}{$\beta_2$} & \multicolumn {1}{c}{$\beta_3$} & \multicolumn {1}{c}{$\beta_4$} & \multicolumn {1}{c}{$\beta_5$} & \multicolumn {1}{c}{$\beta_6$} & \multicolumn {1}{c}{$\beta_7$} \\
      \cmidrule(lr){2-9}
    MGVBP & 0.679 & -1.490 & -0.083 & -0.574 & 0.494 & -0.640 & 0.607 & 0.048 \\
    MGVB  & 0.679 & -1.489 & -0.083 & -0.574 & 0.493 & -0.639 & 0.607 & 0.048 \\
    QBVI  & 0.679 & -1.489 & -0.083 & -0.574 & 0.493 & -0.639 & 0.607 & 0.048 \\
    \midrule
    MGVBP$^{h\text{-func.}}$ & 0.678 & -1.487 & -0.084 & -0.574 & 0.492 & -0.638 & 0.609 & 0.048 \\
    MGVB$^{h\text{-func.}}$ & 0.678 & -1.487 & -0.084 & -0.574 & 0.492 & -0.638 & 0.609 & 0.048 \\
    QBVI$^{h\text{-func.}}$ & 0.678 & -1.487 & -0.084 & -0.574 & 0.492 & -0.638 & 0.609 & 0.048 \\
    \midrule
    MGVBP$^\text{diag.}$ & 0.544 & -1.414 & -0.045 & -0.529 & 0.494 & -0.629 & 0.583 & 0.119 \\
    MGVB$^\text{diag.}$ & 0.542 & -1.413 & -0.044 & -0.529 & 0.493 & -0.629 & 0.583 & 0.121 \\
    QBVI$^\text{diag.}$ & 0.546 & -1.415 & -0.045 & -0.530 & 0.494 & -0.629 & 0.583 & 0.119 \\
    \midrule
    MGVBP$^{h\text{-func. diag}}$ & 0.544 & -1.413 & -0.047 & -0.531 & 0.492 & -0.628 & 0.585 & 0.121 \\
    MGVB$^{h\text{-func. diag}}$ & 0.541 & -1.411 & -0.046 & -0.530 & 0.492 & -0.629 & 0.586 & 0.123 \\
    QBVI$^{h\text{-func. diag}}$ & 0.545 & -1.414 & -0.047 & -0.531 & 0.492 & -0.628 & 0.585 & 0.121 \\
    \midrule
    MCMC  & 0.679 & -1.487 & -0.084 & -0.574 & 0.493 & -0.638 & 0.607 & 0.047 \\
    ML    & 0.679 & -1.476 & -0.085 & -0.571 & 0.485 & -0.625 & 0.599 & 0.045 \\
    \bottomrule
\end{tabular}%
}  
\label{tab:logistic_mu}
\end{table*}

Table \ref{tab:logistic_mu} underlines the unbiasedness of the MGVBP posterior estimates of the variational parameter with respect to the MCMC baseline and the ML target for large samples where the prior is asymptotically negligible.

Similarly, the variational variances of the model parameters in Table \ref{tab:logistic_var} are closely aligned between MGVBP, the other models, and the MCMC and the ML counterparts. 
As well the covariance matrix also closely replicates the one from the MCMC chain; see Table \ref{tab:logistic_cov}. 

The above empirically validates the suitability of the Gaussian approximation, the unbiasedness of the MGVBP posterior estimates of the variational parameter, and its overall consistency in terms of the asymptotic ML covariance matrix.

A diagonal constraint on the covariance harms the performance metrics as a consequence of a different value of the optimized LB as for \ref{tab:logistic_perf}. This results from the different overall levels of the estimated posterior means and variances under this setup (see Table \ref{tab:logistic_mu}, and Table \ref{tab:logistic_var}), and is  
expected considering the considerably smaller overall number of free parameters. However, the variational parameters and the corresponding performance metrics are still aligned between the different optimizers.
As an insight, we comment that the diagonal assumption practically translates the variational Gaussian as the posterior means indeed shift. Indeed, variational methods can be biased with respect to the true posterior distribution \citep{Carbonetto2012Scalable}. This is, after all, justified, thinking that the LB and its gradients, see, e.g., eq. \eqref{eq:grad_S_tran}, depend on the interaction of variational elements concerning both the variational mean and variational covariance elements. I.e., the variational mean under the full case would, in general, differ from that under the diagonal case.

 \begin{table*}
    \centering
    \caption{Variances of the variational approximation on the parameters for the Labour dataset. Entries are multiplied by $10^2$.}
        \scalebox{0.75}{    
\begin{tabular}{lSSSSSSSS}
      & \multicolumn{1}{c}{$\beta_0$} & \multicolumn{1}{c}{$\beta_1$} & \multicolumn{1}{c}{$\beta_2$} & \multicolumn{1}{c}{$\beta_3$} & \multicolumn{1}{c}{$\beta_4$} & \multicolumn{1}{c}{$\beta_5$} & \multicolumn{1}{c}{$\beta_6$} & \multicolumn{1}{c}{$\beta_7$}  \\
\cmidrule{2-9}
    MGVBP & 3.956 & 5.418 & 0.604 & 1.531 & 1.376 & 2.105 & 1.964 & 4.332 \\
    MGVB  & 4.288 & 5.644 & 0.624 & 1.496 & 1.354 & 2.191 & 2.024 & 4.314 \\
    QBVI  & 3.983 & 5.446 & 0.608 & 1.540 & 1.384 & 2.118 & 1.977 & 4.361 \\
    MGVBP$^{h\text{-func.}}$ & 4.071 & 5.377 & 0.596 & 1.463 & 1.284 & 2.057 & 1.972 & 4.389 \\
    MGVB$^{h\text{-func.}}$ & 4.043 & 5.384 & 0.594 & 1.457 & 1.284 & 2.070 & 1.963 & 4.372 \\
    QBVI$^{h\text{-func.}}$ & 4.072 & 5.378 & 0.596 & 1.463 & 1.284 & 2.057 & 1.972 & 4.389 \\
    \midrule
    MGVBP$^\text{diag.}$ & 0.815 & 3.304 & 0.255 & 0.930 & 1.086 & 1.066 & 0.915 & 1.367 \\
    MGVB$^\text{diag.}$ & 0.808 & 3.299 & 0.255 & 0.882 & 1.107 & 1.059 & 0.913 & 1.321 \\
    QBVI$^\text{diag.}$ & 0.817 & 3.311 & 0.255 & 0.932 & 1.088 & 1.068 & 0.917 & 1.371 \\
    MGVBP$^{h\text{-func. diag}}$ & 0.852 & 3.222 & 0.247 & 0.900 & 1.005 & 1.018 & 0.935 & 1.366 \\
    MGVB$^{h\text{-func. diag}}$ & 0.853 & 3.228 & 0.248 & 0.878 & 1.013 & 1.012 & 0.940 & 1.348 \\
    QBVI$^{h\text{-func. diag}}$ & 0.853 & 3.227 & 0.248 & 0.901 & 1.006 & 1.019 & 0.936 & 1.368 \\
    \midrule
    MCMC  & 3.967 & 5.364 & 0.589 & 1.406 & 1.274 & 2.071 & 1.966 & 4.248 \\
    ML    & 4.079 & 5.436 & 0.589 & 1.457 & 1.276 & 2.059 & 1.960 & 4.393 \\
    \bottomrule
\end{tabular}%
\label{tab:logistic_var}
}
\end{table*}

 \begin{table}[ht]
    \centering
    \caption{Variational covariance matrices, MCMC covariance matrix and ML asymptotic covariance matrix. Entries are multiplied by $10^2$.}
\scalebox{0.7}{    
\begin{tabular}{ccSSSSSSSS}&\\
& & \multicolumn{8}{c}{MGVB} \\
\cmidrule{3-10} &   & \multicolumn{1}{c}{$\beta_0$} & \multicolumn{1}{c}{$\beta_1$} & \multicolumn{1}{c}{$\beta_2$} & \multicolumn{1}{c}{$\beta_3$} & \multicolumn{1}{c}{$\beta_4$} & \multicolumn{1}{c}{$\beta_5$} & \multicolumn{1}{c}{$\beta_6$} & \multicolumn{1}{c}{$\beta_7$} \\
\midrule
\multirow{8}[2]{*}{\rotatebox[origin=c]{90}{MGVBP}}
    &$\beta_0$ &       & -1.710 & -0.877 & -0.674 & 0.163 & 0.478 & 0.031 & -2.716 \\
    &$\beta_1$ & -1.458 &       & 0.282 & 1.465 & -0.561 & -0.192 & 0.229 & -0.002 \\
    &$\beta_2$ & -0.790 & 0.218 &       & 0.385 & 0.089 & 0.035 & -0.024 & -0.090 \\
    &$\beta_3$ & -0.712 & 1.406 & 0.406 &       & -0.014 & 0.123 & -0.063 & -0.414 \\
    &$\beta_4$ & 0.135 & -0.395 & 0.100 & 0.046 &       & -0.104 & -0.280 & -0.131 \\
    &$\beta_5$ & 0.228 & -0.136 & 0.031 & 0.082 & -0.111 &       & -1.366 & -0.743 \\
    &$\beta_6$ & 0.235 & 0.177 & -0.057 & -0.176 & -0.313 & -1.305 &       & -0.039 \\
    &$\beta_7$ & -2.584 & -0.090 & -0.124 & -0.389 & -0.199 & -0.561 & -0.122 &  \\
    \bottomrule\\
\end{tabular}%
}

\scalebox{0.7}{    
\begin{tabular}{ccSSSSSSSS}
& & \multicolumn{8}{c}{ML} \\
\cmidrule{3-10} &   & \multicolumn{1}{c}{$\beta_0$} & \multicolumn{1}{c}{$\beta_1$} & \multicolumn{1}{c}{$\beta_2$} & \multicolumn{1}{c}{$\beta_3$} & \multicolumn{1}{c}{$\beta_4$} & \multicolumn{1}{c}{$\beta_5$} & \multicolumn{1}{c}{$\beta_6$} & \multicolumn{1}{c}{$\beta_7$} \\
\midrule
\multirow{8}[2]{*}{\rotatebox[origin=c]{90}{MCMC}}
    &$\beta_0$ &       & -1.578 & -0.800 & -0.684 & 0.067 & 0.318 & 0.194 & -2.701 \\
    &$\beta_1$ & -1.510 &       & 0.274 & 1.378 & -0.422 & -0.027 & 0.074 & -0.084 \\
    &$\beta_2$ & -0.791 & 0.249 &       & 0.393 & 0.102 & 0.024 & -0.058 & -0.093 \\
    &$\beta_3$ & -0.654 & 1.307 & 0.381 &       & 0.067 & 0.117 & -0.164 & -0.352 \\
    &$\beta_4$ & 0.047 & -0.414 & 0.111 & 0.072 &       & -0.172 & -0.265 & -0.109 \\
    &$\beta_5$ & 0.334 & -0.049 & 0.006 & 0.089 & -0.177 &       & -1.286 & -0.613 \\
    &$\beta_6$ & 0.172 & 0.111 & -0.036 & -0.137 & -0.266 & -1.282 &       & -0.097 \\
    &$\beta_7$ & -2.603 & -0.085 & -0.097 & -0.349 & -0.119 & -0.590 & -0.102 &  \\
    \bottomrule
\end{tabular}%
}
\label{tab:logistic_cov}
\end{table}

\begin{figure}[!ht]
    \centering
    \includegraphics[trim={2.9cm 0.2cm 1.4cm 0cm},clip,width = 1.05\textwidth]{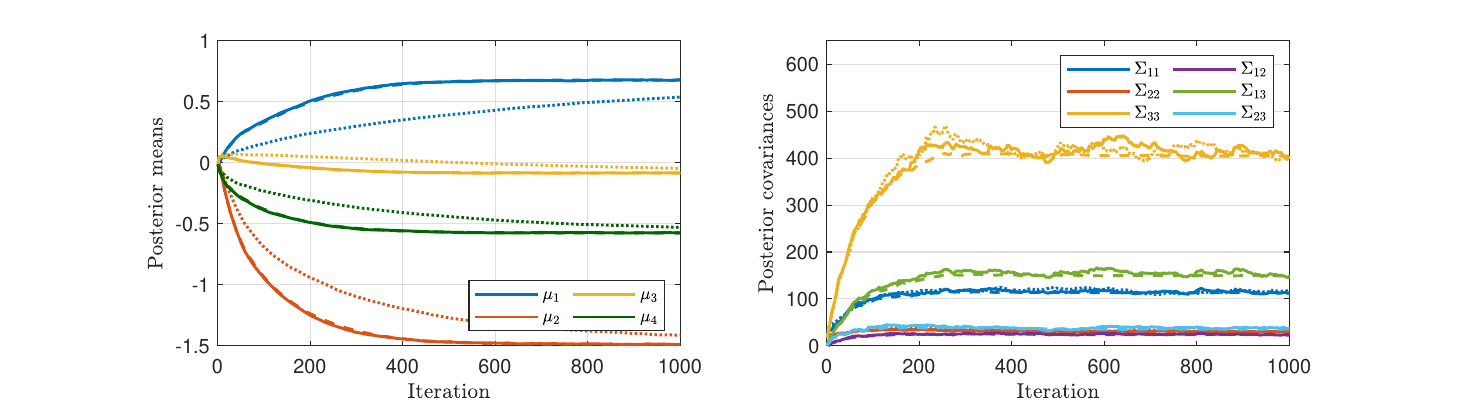}
    \caption{Parameter learning across the iterations for the MGVBP algorithm on the Labour dataset for some selected variational parameters. Dotted lines correspond to the diagonal case, dashed lines to the use of the $h$-function gradient estimator.}
    \label{fig:Labour_PostMeanCov}
\end{figure}

Furthermore, Figure \ref{fig:Labour_PostMeanCov} shows that the variational mean and covariance learning dynamics are smooth and steady, without wigglings or anomalies. At the same time, using the $h$-function estimator stabilizes the learning process, but has a minor effect on the maximized LB (Tables \ref{tab:logistic_perf}, \ref{tab:logistic_mu}, \ref{tab:logistic_var}).
Indeed, the adoption of the gradient estimator based on the $h$-function has an overall minimal impact on the optimized results. In fact, the $h$-function alternative impacts only the variance of the gradient estimator without affecting the expected direction of the steps in the SGD update.

Table \ref{tab:logistic_S} in Appendix \ref{app:experiments}, addressed the impact the number of MC samples $S$ has on the posterior mean, performance measures, and the optimized LB. is minor for both the training and test phases (Table \ref{tab:logistic_S}). We conclude that about 20 samples are sufficient (also for complex models such as the following neural network) and are a feasible compromise between the quality of the MC approximation of the LB gradient and computational efficiency (also in the following experiments). Our experiments are, however, oriented toward discussing the improvements and advantages of MGVBP over MGVB. To this end, we require a precise estimation of the LB and thus employ a much higher value of $S$, see Table \ref{tab:hyperparameters}.

We additionally train a (bilinear) neural network with a TABL layer \citep{tran_temporal_2019}. The bottom panel of Table \ref{tab:tabl} reports the estimation results on the TABL neural network, referred to as the LOB experiment. The prediction of high-frequency mid-price movements in limit-order book markets is known to be a challenging task, where Bayesian techniques are particularly relevant for implementing risk-aware trading strategies \citep{magris2022bayesian}, and where the TABL architecture is very effective.
Training MGVB on such a machine learning model empirically appears unstable and generally unfeasible, both with our own and the original implementation of \cite{tran_variational_2021}. Indeed we are unaware of any MGVB application for typical machine learning models.
We adopt different baselines.
Because of the inability of QBVI to grant the positive definite constraint, we train it under a diagonal assumption. Furthermore, we include the state-of-the-art VOGN optimizer \citep{osawa_practical_2019}. VOGN also assumes a diagonal Gaussian posterior and relies on natural gradient computations employing models' (per-sample) gradients and an approximate Hessian. Its second-order nature closely resembles the standard ADAM optimizer, included as a non-Bayesian baseline, included as a non-Bayesian baseline. 

\begin{figure}
    \centering
    \includegraphics[trim={1cm 0.3cm 0.8cm 0.5cm},clip,width = 0.75\textwidth]{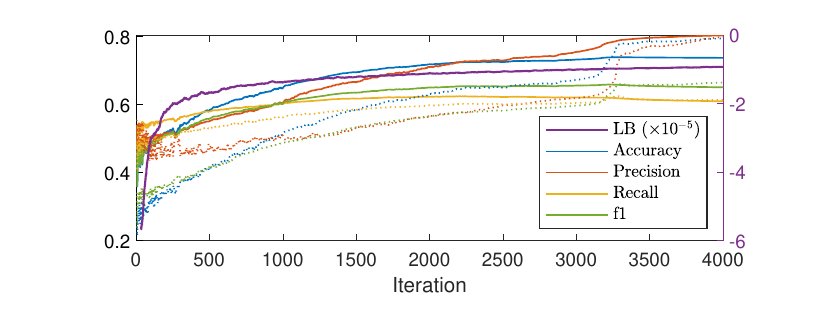}  
    \caption{Lower bound (right axis) and performance measure cross iterations (left axis) for the LOB experiment on the train set (continuous lines) test set (dotted lines) for MGVBP.}
    \label{fig:tabl}
\end{figure}

The problem is highly non-convex: indeed, we observe the algorithms converging at different optima and heterogeneous performance measures. Yet, MGVBP clearly outperforms the other Bayesian alternatives on both the training and test set, promoting its use beyond standard statistical models. Concerning the MCMC estimates, we observe ubiquitous evidence of multi-modality, skewness, and asymmetry in the margins with an underlying non-Gaussian copula, that any fixed-form Gaussian VI can not, by construction, deal with. However, the peak performance obtained with the Gaussian Assumption under MGVBP suggests a main mode where most of the density is concentrated,  captured by the VI approximation. The evolution of the learning process for the LB and performance metrics in Figure \ref{fig:tabl} remarks a very stable behavior of the stochastic LB estimate ($N_s = 20$ only), along with a consistent and resolved improvement of the performance measures since early epochs. The performance under MGVBP is also remarkably improved on the test set.

\begin{table*}
\centering
  \caption{Performance measures for the LOB experiment.}
  \scalebox{0.65}{ 
    \begin{tabular}{lccccccccc}
          & \multicolumn{5}{c}{Train}  & \multicolumn{4}{c}{Test} \\ \cmidrule(lr){2-6} \cmidrule(lr){7-10}
          & $\LB^\star$ & Accuracy & Precision & Recall & f1    &  Accuracy & Precision & Recall & f1 \\
    \midrule
    MGVBP & \;\;-91948.593 & 0.738  &  0.804   & 0.610  &  0.651  & 0.795 &   0.798  &  0.614  &  0.665\\
    QBVI \textsuperscript{(diag.)}  & -104261.698 & 0.717 & 0.697 & 0.637 & 0.658 &  0.686 & 0.650 & 0.605 & 0.603\\
    VOGN \textsuperscript{(diag.)}  &  -100791.406& 0.673& 0.631& 0.629& 0.630 &  0.758& 0.677& 0.637& 0.652\\
    MCMC  & & 0.689&0.773&0.536&0.567 &0.761&0.779&0.536&0.585\\
    ADAM \textsuperscript{(non Bayesian)}  &&&&&& 0.755& 0.668& 0.640& 0.652\\
    \bottomrule
    \end{tabular}%
    }
  \label{tab:tabl}%
\end{table*}%

\subsection{Regression tasks}

\begin{table}[htbp]
    \centering
    \caption{Regression task. Parameters' estimates and performance measures.}
    \scalebox{0.63}{ 
    \begin{tabular}{lcccccccccc}
     & &&&&\multicolumn{4}{c}{Train}  & \multicolumn{2}{c}{Test} \\ \cmidrule(lr){6-9} \cmidrule(lr){10-11}
          & $\omega$ & $\alpha$ & $\gamma$ & $\beta$ & $\LB^\star$ & $p\br{y\vert \mu^\star}$ & MSE\textsubscript{rv} & QLIKE\textsubscript{rv} &  MSE\textsubscript{rv} & QLIKE\textsubscript{rv} \\
\cmidrule{2-11}    \textit{\textbf{GARCH(1,1)}} &       &       &       &       &             &       &       &       &       &  \\
    MGVBP & 0.043 & 0.230 &       & 0.737 & -2027.489 & -2017.633 & 5.124 & 0.531 &  2.564 & 0.566 \\
    MGVB  & 0.043 & 0.229 &       & 0.737 & -2027.493 & -2017.626 & 5.113 & 0.531 &  2.561 & 0.566 \\
    QBVI  & 0.043 & 0.229 &       & 0.737 & -2027.493 & -2017.626 & 5.114 & 0.531 & 2.561 & 0.566 \\
    MCMC  & 0.042 & 0.231 &       & 0.738 &       & -2017.713 & 5.189 & 0.531 &  2.565 & 0.566 \\
    ML    & 0.042 & 0.226 &       & 0.739 &       & -2017.593 & 5.073 & 0.531 & 2.531 & 0.566 \\
    \midrule
    \textit{\textbf{GJR(1,1) }} &       &       &       &       &       &       &              &       &       &  \\
    MGVBP & 0.043 & 0.108 & 0.294 & 0.722 & -2003.542 & -1990.907 & 7.039 & 0.310 &  1.556 & 0.346 \\
    MGVB  & 0.044 & 0.108 & 0.292 & 0.722 & -2003.550 & -1990.896 & 6.969 & 0.310 &  1.551 & 0.347 \\
    QBVI  & 0.044 & 0.108 & 0.292 & 0.722 & -2003.550 & -1990.897 & 6.971 & 0.310 & 1.551 & 0.347 \\
    MCMC  & 0.043 & 0.108 & 0.298 & 0.724 &       & -1991.095 & 7.318 & 0.309 &  1.576 & 0.344 \\
    ML    & 0.042 & 0.108 & 0.291 & 0.723 &       & -1990.853 & 6.940 & 0.311 & 1.548 & 0.349 \\
    \midrule
    \textit{\textbf{HAR}} & $\beta_0$ & $\beta_1$ & $\beta_2$ & $\beta_3$ &              &       & $10^2\times$QLIKE\textsubscript{rv} &       &       & $10^2\times$QLIKE\textsubscript{rv} \\
\cmidrule{2-11}    MGVBP & 1.032 & 0.489 & 0.420 & -0.009 & -5080.191 & -5058.113 & 24.193 & 5.324 &  18.861 & 5.712 \\
    MGVB  & 1.019 & 0.485 & 0.426 & -0.011 & -5082.123 & -5058.167 & 24.194 & 5.328 &  18.879 & 5.720 \\
    QBVI$^\text{diag.}$ & 1.036 & 0.485 & 0.423 & -0.007 & -5083.253 & -5058.143 & 24.193 & 5.326 &  18.862 & 5.710 \\
    MCMC  & 1.075 & 0.488 & 0.421 & -0.012 &       & -5058.090 & 24.192 & 5.328 & 18.858 & 5.706 \\
    ML    & 1.080 & 0.488 & 0.421 & -0.012 &       & -5058.086 & 24.192 & 5.328 & 18.857 & 5.706 \\
    \bottomrule
    \end{tabular}%
    }
  \label{tab:regression}%
\end{table}%

The GARCH-related models in the lower panels of Table \ref{tab:regression} address the predictive ability of MGVBP under non-standard and non-convex likelihood functions.
As performance measures, we adopt the Mean Squared Error (MSE) and the QLIKE loss \citep{patton2011volatility} computed with respect to the 5-minute sub-sampled realized volatility \citep{zhang2005tale} as a robust proxy for daily compared to squared daily returns. Additional information and extended results appear in Appendix \ref{app:volatility}, also involving performance measures computed with respect to the squared daily returns.

\begin{figure}[!ht]
    \centering
    \includegraphics[trim={2cm 0cm 1cm 0cm}, clip, width = 1.05\textwidth]{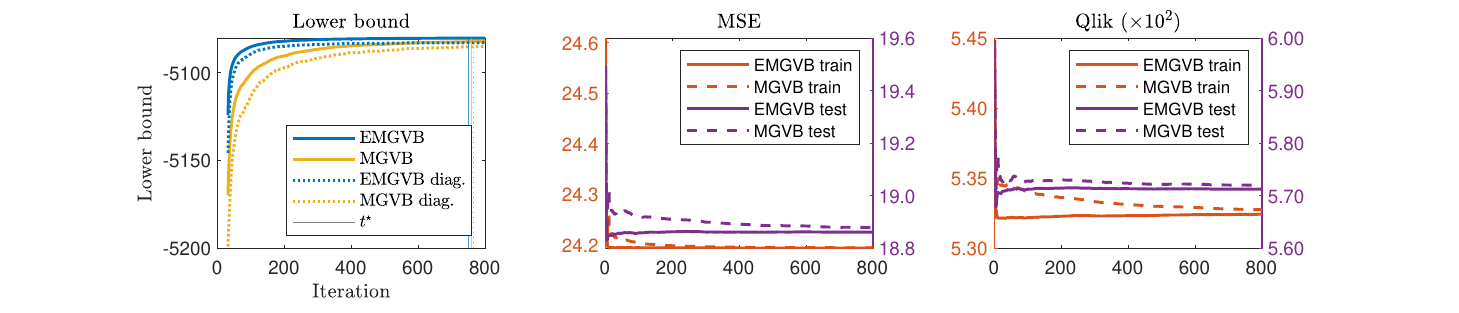}
    \includegraphics[trim={2cm 0cm 1cm 0cm}, clip, width = 1.05\textwidth]{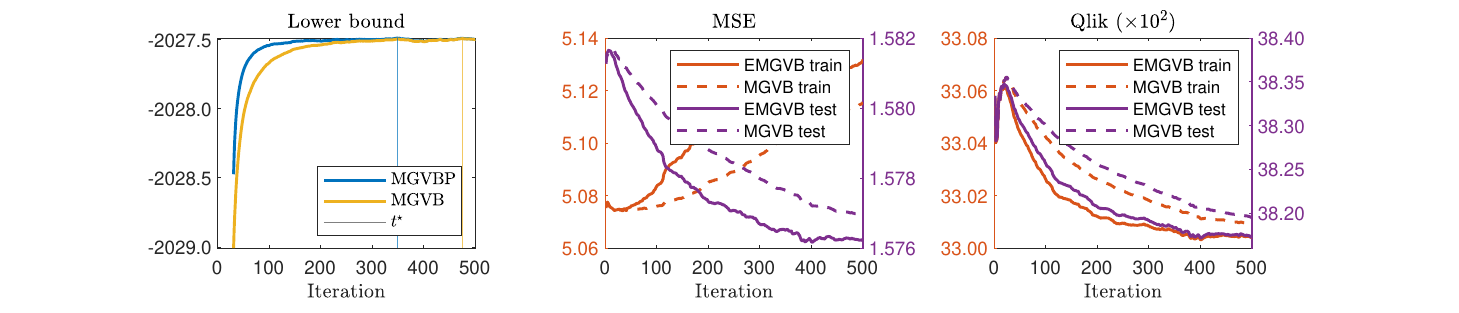}
    \includegraphics[trim={2cm 0cm 1cm 0cm}, clip, width = 1.05\textwidth]{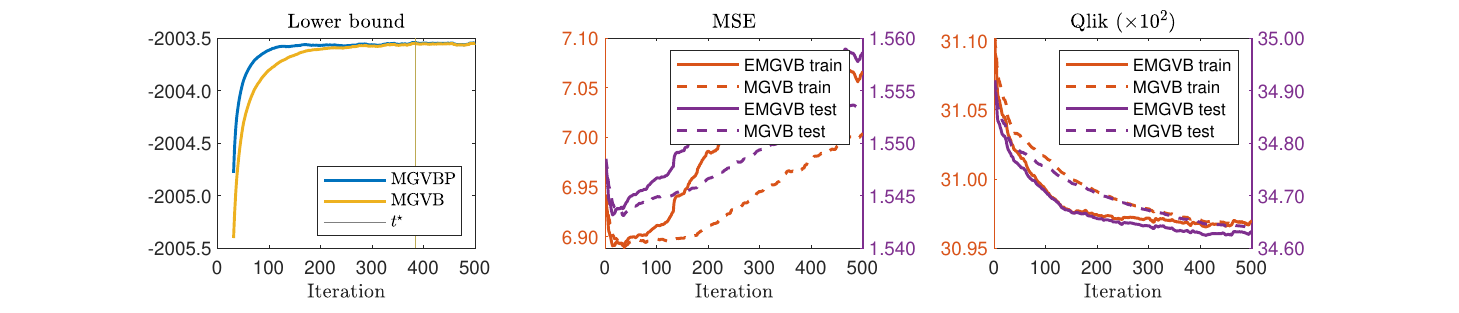}
    \caption{Progression of the lower bound optimization, dynamics of the MSE and QLIKE, both computed with respect to RV5\textsubscript{ss}, across the iterations for the HAR model (top row), GARCH(1,1) model (middle row) and GJR(1,1) (bottom row). Vertical lines denote the iteration $t^\star$, for which the line style and color match the one of the corresponding optimizer. For the GARCH and GJR, see Appendix \ref{app:further_cons} for the corresponding plots of the parameters' posterior margins.}
    \label{fig:regression}
\end{figure}

As for the logistic regression experiments, we observe that all the algorithms approach the very same ML optimum, indicating that both MGVBP and MGVB move towards the same LB maximum. Furthermore, all variational approximations are well-aligned with the MCMC and ML estimates. Thus, the estimates, performance metrics, and value of the optimized LB are similar across the optimizers: they all converge to the minimum but in a {\it qualitatively} different way. Indeed, in \ref{fig:regression}, the LB improvement across the iterations is steeper for MGVBP and also for a diagonal implementation. With respect to the dynamics of the performance metrics computed on the test depicted on the second and third columns of the plots in Figure  \ref{fig:regression}, we observed that MGVBP dominates, both in terms of MSE and QLIKE the MGVB baseline for the HAR model. On the other hand, for the GARCH family models, whereas MGVBP clearly outperforms MGVB in terms of the QLIKE loss both on the training and test set, this dominance is not so evident for the MSE. 
In fact, there is no direct link between the dynamics of the LB and that of the performance measures, i.e., there are no trivial guarantees that, e.g., as the LB is maximized, the MSE is minimized.
In fact, while this is the case for the HAR model, for the GARCH modes, we observe an increase of the MSE across the iterations, despite the fact that the LB approaches the optimum. The same hold for the test MSE: for the GARCH(1,1) it decreases across the iterations, for the GJR(1,1) not. This should not be surprising as the optimization objective is the overall minimization of the KL divergence between the true posterior and the variational one, achieved via LB maximization, and not the minimization of the MSE, Q-link, or other performance metrics.
Figure \ref{fig:regression} validates this interpretation and reminds us that VI has to be interpreted for what it is: a solution for approximating the posterior, not a performance-metric minimizing tool, as, e.g., ordinary least squares (w.r.t. the squared loss) or ML (w.r.t. model negative loglikelihood). Consequently, the reference measure for an actual comparison of the VI algorithms should be the value of the optimized lower bound and its dynamics.
The additional results in Appendix \ref{app:volatility} confirm the improved ability of MGVBP in optimizing the LB and the overall alignment between the solution obtained with MGVB, MCMC, and ML on further GARCH-type models. See tables \ref{tab:vol_est},\ref{tab:vol_est_EGARCH},\ref{tab:vol_est_FIGARCH_HAR}, and the Figures \ref{fig:garch},\ref{fig:gjr},\ref{fig:figarch}, comparing the posterior marginals obtained via VI and MCMC.

\subsection{Non-differentiable model}
Table \ref{tab:non_diff} reports the estimation results for a non-differentiable model, where, e.g., the reparametrization trick and methods generally relying on gradients and automatic differentiation are inapplicable. We adopt the reconstruction benchmark
problem in \citep{lyu2019black}:
\begin{align*}
  y_i &= f\br{\bm{x}_i} + \varepsilon_i\text{,}\qquad \varepsilon_i \sim  \N\br{0,\sigma^2}\text{,}\\
  f\br{\bm{x}_i} &= \vert \vert \text{sign} \br{\bm{x}_i-0.5} -\beta\vert \vert _2^2 - \vert \vert \text{sign}\br{\beta} - \beta \vert \vert _2^2\text{,} \qquad 
  \bm{x}_i \in \left\lbrace 0,1 \right\rbrace^d\text{.}
\end{align*}
We simulate $i=1,\dots,5000$ observations for the above model with $d = 20$, $\beta \sim \text{Uniform}\br{0,1}$, $p=0.5$, $\sigma^2 = 1$, and $x$ sampled from $d$ independent Bernoulli distributions. MGVBP and MGVB are both implemented over the $h$-function estimator and for a full specification of the variational covariance matrix. For reference, the initial value of the lower bound for all the experiments is $-11591.339$, and we include the mean absolute deviation (MAD). Details on the hyperparameters are provided in Table \ref{tab:hyperparameters}.

\begin{table}[htbp]
  \centering
  \caption{Results for the non-differentiable model.}
  \scalebox{0.8}{
    \begin{tabular}{lccccccc}
    & \multicolumn{4}{c}{Train}     & \multicolumn{3}{c}{Test} \\
    \cmidrule(lr){2-5}   \cmidrule(lr){6-8}  
    & $\LB^\star$ & $p(y\vert\bm{\mu}^\star)$ & MSE   & MAD   & $p(y\vert\bm{\mu}^\star)$ & MSE   & MAD \\
    \cmidrule{2-8}    MGVB  & -2901.67 & -2848.05 & 1.020  & 0.800   & -707.451 & 1.033 & 0.798 \\
    MGVBP & -2898.86 & -2845.41 & 1.018 & 0.798 & -705.945 & 1.027 & 0.794 \\
    QBVI\textsuperscript{diag.} & -2904.48 & -2846.67 & 1.019 & 0.798 & -706.78 & 1.03  & 0.796 \\
    \bottomrule
    \end{tabular}%
    }
  \label{tab:non_diff}%
\end{table}%

From Table \ref{tab:non_diff}, we confirm the overall accordance of the three black box approaches considered in this work and the improved performance of MGVBP also for the above non-differentiable model.

\section{Conclusion}\label{sec:conclusion}
We propose an algorithm based on manifold optimization to guarantee the positive-definite constraint on the covariance matrix of the Gaussian variational posterior. Extending the baseline method of \citep{tran_variational_2021}, we exploit the computational advantage of the Gaussian parametrization in terms of the precision matrix and employ simple and computationally convenient analytical solutions for the natural gradients. Furthermore, we provide a theoretically consistent setup justifying the use of the SPD-manifold retraction form. Our MGVBP black-box optimizer results in a ready-to-use solution for VI, scalable to structured covariance matrices, that can take advantage of alternative forms of the stochastic gradient estimator, control variates, and momentum. 
We show our solution's feasibility on many statistical, econometric, and ML models over different baseline optimizers. Our results align with sampling methods and highlight the advantages of the suggested approach over state-of-the-art baselines in terms of convergence and performance metrics. 
Future research may investigate the applicability of our approach to a broader set of variational distributions, and the use of the reparametrization trick in place of black-box gradients.

\section*{Acknowledgments}
This project has received funding from the European Union’s Horizon 2020 research and innovation programme under the Marie Sk\l odowska-Curie grant agreement No. 890690, and the Independent Research Fund Denmark project DISPA (project No. 9041-00004)

\bibliographystyle{APA}
\bibliography{biblio}

\appendix
\section{Block-diagonal implementation}\label{app:further_cons}
{\setstretch{1.0}
\begin{algorithm}[H]
\centering
\caption{MGVBP for a block-diagonal covariance matrix (prior with zero-mean and covariance matrix $\tau^{-1} I$)}\label{alg:mean_field}
\begin{algorithmic}[1]
\STATE \text{Set hyper-parameters: }$ 0<\beta,\omega< 1$, $S$
\STATE \text{Set the type of gradient estimator, i.e. function $\log f\br{\bth_s}$}
\STATE \text{Set prior $p\br{\bth;\bm{0},\tau I}$, likelihood $p\br{\by\vert\bth}$, and initial values $\bmu$, $\iS$}
\STATE $t=1$, $\text{Stop} = \texttt{false}$
\STATE \text{Generate: } $\bth_s = \sbr{\bth_{s_1},\dots,\bth_{s_h}}$, \quad $\bth_{s_i} \sim q_{\bmu_i,\S_i}$, $s = 1,\dots,S$, $i = 1,\dots,h$
\FOR{$i = 1,\dots, h$}
\STATE \text{Compute: } $\hat{g}_{\bmu_i} = \S_i \hat{\nabla}_{\bmu_i}\LB$,\quad $\hat{g}_{\iS_i} = -\hat{\nabla}_{\S_i}\LB$
\STATE $\bm{m}_{\bmu_i} = \hat{g}_{\bmu_i}$,\quad $\bm{m}_{\iS_i} = \hat{g}_{\iS_i}$ 
\ENDFOR

\WHILE{$\text{Stop} = \texttt{false}$}
\FOR{$i = 1,\dots, h$}
\STATE $\bmu_i = \bmu_i + \beta \bm{m}_{\bmu_i} $ 
\STATE $\iS_{\text{old},i} = \iS_i$,\quad $\iS_i = R_{\iS_{\text{old},i}}(\beta \bm{m}_{\iS_i}) $
\ENDFOR
\STATE \text{Generate: } $\bth_s = \sbr{\bth_{s_1},\dots,\bth_{s_h}}$, \quad $\bth_{s_i} \sim q_{\bmu_i,\S_i}$, $s = 1\dots S$, $i = 1,\dots,h$
\STATE \text{Set: } $\log q_s = 0$, $s = 1,\dots,S$
\FOR{$i = 1,\dots, h$}
\STATE \text{Compute: } $\hat{g}_{\bmu_i}$, $\hat{g}_{\iS_i}$  
\STATE $\bm{m}_{\bmu_i} = \omega \bm{m}_{\bmu_i} + \br{1-\omega} \hat{g}_{\bmu_i}$  
\STATE $\bm{m}_{\iS_i} = \mathcal{T}_{\iS_{\text{old},i} \rightarrow \iS_i}(\bm{m}_{\iS_i})+\br{1-\omega} \hat{g}_{\iS_i}$ 
\STATE $\log q_s = \log q_s +\log q_{\bmu_i,\S_i}\br{\bth_{s_i}}$
\ENDFOR
\STATE $\hat{\LB} = \frac{1}{S}\sum_{s=1}^S \sbr{\log p \br{\bth_s} + \log p\br{\by\vert \bth_s} + \log q_s}$
\STATE $t = t+1$, $\text{Stop} =  f_\text{exit}\br{t,\dots}$  

\ENDWHILE
\end{algorithmic}
\end{algorithm}
}
\FloatBarrier

\section{Experiments}\label{app:experiments}
\subsection{Additional results for the labour data} \label{app:res:labour}

\begin{table*}
    \centering 
        \caption{Estimated parameters and performance measures on the Labour dataset for MGVBP (full-posterior) for different sizes of the number of MC draws for the estimation of the stochastic gradients $S$. $t$ refers to the run-time per iteration (in milliseconds), $\mathcal{L}(\bth_0)$ to the LB evaluated at the initial parameters. For each $S$, a common random seed is used.}
    \scalebox{0.7}{    
\begin{tabular}{cSSSSSSSSSS}
\multicolumn {1}{c}{$S$}   & \multicolumn {1}{c}{$t$}   & \multicolumn {1}{c}{$\beta_0$} & \multicolumn {1}{c}{$\beta_1$} & \multicolumn {1}{c}{$\beta_2$} & \multicolumn {1}{c}{$\beta_3$} & \multicolumn {1}{c}{$\beta_4$} & \multicolumn {1}{c}{$\beta_5$} & \multicolumn {1}{c}{$\beta_6$} & \multicolumn {1}{c}{$\beta_7$} & \multicolumn {1}{c}{$\LB\br{\bm{\zeta}^c_0}$} \\
    \toprule
    10    & 0.002 & 0.674 & -1.478 & -0.080 & -0.569 & 0.500 & -0.634 & 0.593 & 0.041 & -430.216 \\
    20    & 0.002 & 0.680 & -1.490 & -0.083 & -0.573 & 0.491 & -0.637 & 0.602 & 0.044 & -429.626 \\
    30    & 0.003 & 0.680 & -1.491 & -0.083 & -0.573 & 0.492 & -0.635 & 0.602 & 0.045 & -434.863 \\
    50    & 0.004 & 0.679 & -1.489 & -0.083 & -0.573 & 0.495 & -0.639 & 0.607 & 0.045 & -435.890 \\
    75    & 0.006 & 0.679 & -1.489 & -0.083 & -0.574 & 0.493 & -0.639 & 0.607 & 0.048 & -436.282 \\
    100   & 0.008 & 0.679 & -1.490 & -0.084 & -0.575 & 0.494 & -0.639 & 0.608 & 0.049 & -436.924 \\
    150   & 0.011 & 0.679 & -1.488 & -0.085 & -0.575 & 0.493 & -0.639 & 0.609 & 0.048 & -436.759 \\
    200   & 0.013 & 0.680 & -1.489 & -0.085 & -0.575 & 0.493 & -0.638 & 0.609 & 0.047 & -437.036 \\
    300   & 0.020 & 0.679 & -1.487 & -0.084 & -0.575 & 0.493 & -0.638 & 0.609 & 0.047 & -436.529 \\
    \bottomrule
\end{tabular}%
}
  \scalebox{0.62}{
    \begin{tabular}{cccccccccc}
          & \multicolumn{5}{c}{Train}             & \multicolumn{4}{c}{Train} \\
\cmidrule(lr){2-6} \cmidrule(lr){7-10}
$S$   & $\LB^\star$ & Accuracy & Precision & Recall & f1 & Accuracy & Precision & Recall & f1 \\
    \midrule
    10    & -356.687 & 0.711 & 0.710 & 0.701 & 0.706 & 0.698 & 0.679 & 0.674 & 0.676 \\
    20    & -356.667 & 0.713 & 0.712 & 0.703 & 0.708 & 0.698 & 0.679 & 0.674 & 0.676 \\
    30    & -356.653 & 0.713 & 0.712 & 0.703 & 0.708 & 0.693 & 0.673 & 0.667 & 0.670 \\
    50    & -356.645 & 0.713 & 0.712 & 0.703 & 0.708 & 0.698 & 0.679 & 0.674 & 0.676 \\
    75    & -356.642 & 0.713 & 0.712 & 0.703 & 0.708 & 0.698 & 0.679 & 0.674 & 0.676 \\
    100   & -356.640 & 0.711 & 0.710 & 0.701 & 0.706 & 0.698 & 0.679 & 0.674 & 0.676 \\
    150   & -356.640 & 0.711 & 0.710 & 0.701 & 0.706 & 0.698 & 0.679 & 0.674 & 0.676 \\
    200   & -356.639 & 0.711 & 0.710 & 0.701 & 0.706 & 0.698 & 0.679 & 0.674 & 0.676 \\
    300   & -356.639 & 0.711 & 0.710 & 0.701 & 0.706 & 0.698 & 0.679 & 0.674 & 0.676 \\
    \bottomrule
    \end{tabular}%
    }
    \label{tab:logistic_S}
\end{table*}

\FloatBarrier
\subsection{Volatility models} \label{app:volatility}
Our second set of experiments involves the estimation of several GARCH-family volatility models. The models in Tables \ref{tab:vol_est}, \ref{tab:vol_est_EGARCH}, \ref{tab:vol_est_FIGARCH_HAR} differ for the number of estimated parameters, the form of the likelihood function (which can be quite complex as for the FIGARCH models), and constraints imposed on the parameters. Besides the GARCH-type models, we include the well-known linear HAR model for realized volatility \citep{corsi2009simple}.  We performed a preliminary study for retaining only relevant models, e.g., we observed that for a GARCH(1,0,2) $\beta_2$ is not significant, so we trained a GARCH(1,0,1), or that the autoregressive coefficient of the squared innovations is always significant only at lag one, so we did not consider further lags for $\alpha$. For $\alpha, \beta, \gamma$, we restricted the search up to lag 2. 
Except for HAR's parameter $\beta_3$, all the parameters of all the models are statistically significant under standard ML at $5\%$. Note that the aim of this experiment is to perform VI to the above class of models, not to discuss their empirical performance or forecasting ability. For the reader unfamiliar with the above (standard) econometric models, discussion, and notation we refer, e.g., to the accessible introduction of \citep{terasvirta2009introduction}.

We report the values of the smoothed lower bound computed at the optimized parameter $\LB\br{\bz^\star}$, the model's log-likelihood in the estimates posterior parameter $p\br{\by\vert \bth^\star}$ and the MSE and QLIKE between the fitted values and squared daily returns and subsampled realized variances, used as a volatility proxy. Details on the data and hyperparameters are provided in Appendix \ref{app:exp:hyperp}.

\begin{figure}
\centering
    \includegraphics[trim={2cm 0 2cm 0cm},clip,width = .8\textwidth]{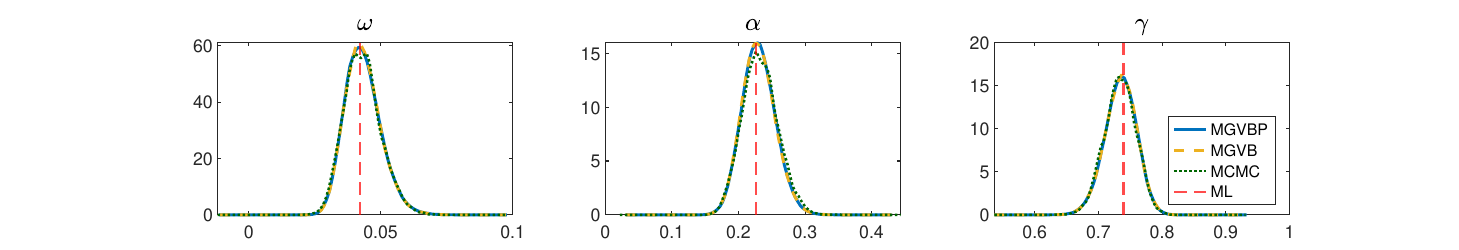}
    \includegraphics[trim={2cm 0 2cm 0cm},clip,width = .8\textwidth]{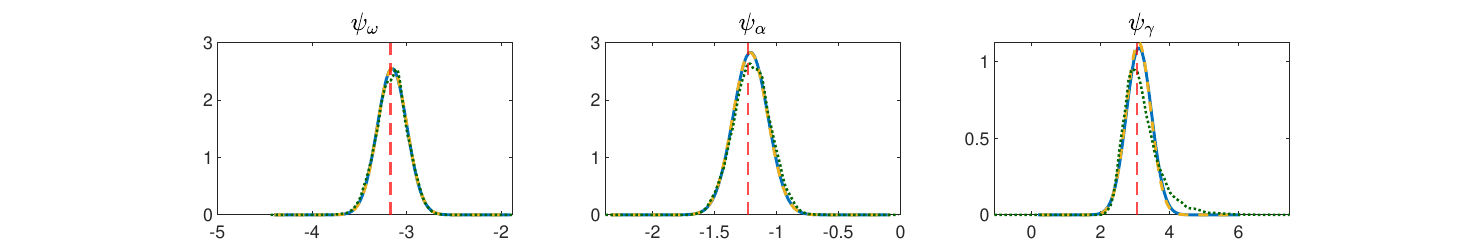}
    \caption{GARCH(1,1) model. MCMC and variational marginals in for the unconstrained parameters $(\psi_\omega,\psi_\alpha,\psi_\gamma,\psi_\beta)$ (top row) and constrained parameters $(\omega,\alpha,\gamma,\beta)$ (bottom row).}
    \label{fig:garch}
\end{figure}

\begin{figure}
\centering
    \includegraphics[trim={2cm 0 2cm 0cm},clip,width = .8\textwidth]{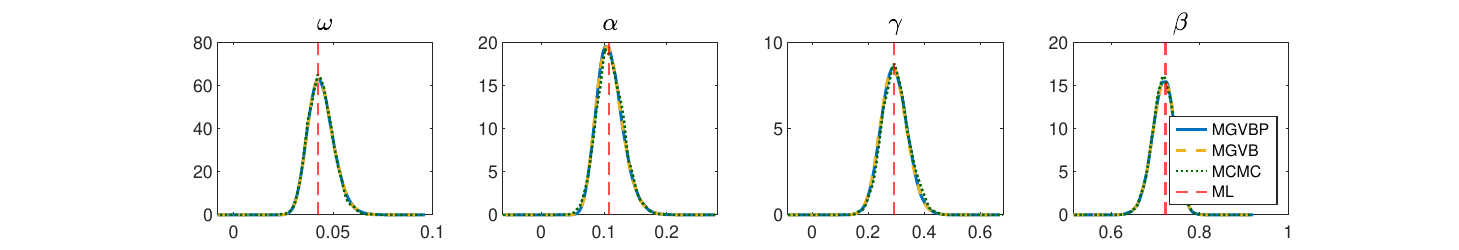}
    \includegraphics[trim={2cm 0 2cm 0cm},clip,width = .8\textwidth]{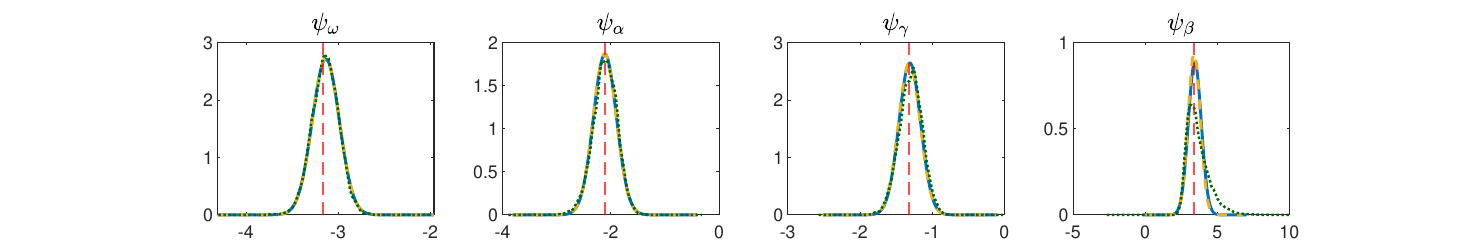}
    \caption{GJR(1,1,1) model. MCMC and variational marginals in for the unconstrained parameters (top row) and constrained parameters (bottom row).}
    \label{fig:gjr}
\end{figure}

\begin{figure}
\centering
    \includegraphics[trim={2cm 0 2cm 0cm},clip,width = .8\textwidth]{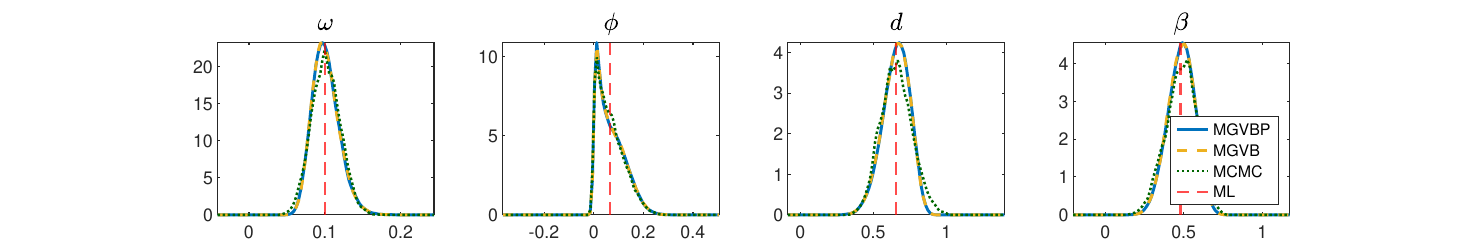}
    \includegraphics[trim={2cm 0 2cm 0cm},clip,width = .8\textwidth]{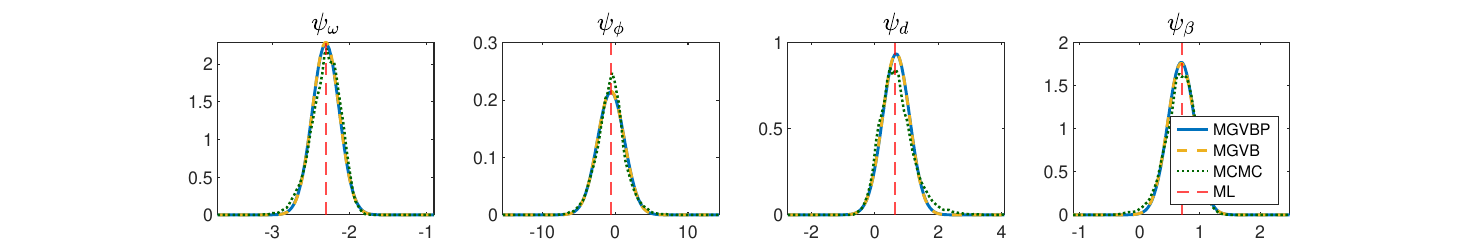}
    \caption{FIGARCH(1,d,1) model. MCMC and variational marginals in for the unconstrained parameters (top row) and constrained parameters (bottom row).}
    \label{fig:figarch}
\end{figure}

A visual inspection of the marginal densities in Figures \ref{fig:garch},\ref{fig:gjr},\ref{fig:figarch}, reveals that, in general, both MGVBP and MGVB perform quite well compared to MCMC sampling and that the variational Gaussian assumption is feasible for all the volatility models. Note that the skews observed in the figures are due to the parameter transformation: VI is applied on the unconstrained parameters $(\psi_\omega, \psi_\phi)$ and such variational Gaussians are back-transformed on the original constrained parameter space where the distributions are generally no longer Gaussian.

\begin{table}[htbp]
  \centering
    \caption{Estimation results and performance metrics for some GARCH variants.}
  \scalebox{0.5}{
    \begin{tabular}{lcccccccccccccccc}
          &       &       &       &       &       & \multicolumn{6}{c}{Train}                     & \multicolumn{5}{c}{Test} \\
          \cmidrule(lr){7-12} \cmidrule(lr){13-17}      
          & $\omega$ & $\alpha$ & $\gamma$ & $\beta_1$ & $\beta_2$ & $\LB^\star$ & $p\br{y\vert \bmu^\star}$ & MSE\textsubscript{r} & MSE\textsubscript{rv} & QLIKE\textsubscript{r} & QLIKE\textsubscript{rv} & $p\br{y\vert \theta^\star}$ & MSE\textsubscript{r} & MSE\textsubscript{rv} & QLIKE\textsubscript{r} & QLIKE\textsubscript{rv} \\
\cmidrule{2-17}          & \multicolumn{16}{c}{ARCH} \\
    MGVBP & 0.519 & 0.640 &       &       &       & -2262.243 & -2256.136 & 29.659 & 8.164 & 1.918 & 0.531 & -665.516 & 6.575 & 2.564 & 1.685 & 0.566 \\
    MGVB  & 0.520 & 0.639 &       &       &       & -2262.244 & -2256.133 & 29.647 & 8.146 & 1.918 & 0.531 & -665.499 & 6.568 & 2.561 & 1.685 & 0.566 \\
    QBVI  & 0.520 & 0.639 &       &       &       & -2262.244 & -2256.133 & 29.647 & 8.146 & 1.918 & 0.531 & -665.498 & 6.568 & 2.561 & 1.685 & 0.566 \\
    MCMC  & 0.519 & 0.640 &       &       &       &       & -2256.136 & 29.660 & 8.165 & 1.918 & 0.531 & -665.537 & 6.575 & 2.565 & 1.685 & 0.566 \\
    ML    & 0.520 & 0.628 &       &       &       &       & -2256.116 & 29.538 & 7.993 & 1.918 & 0.531 & -665.486 & 6.511 & 2.531 & 1.685 & 0.566 \\
    \midrule
          & \multicolumn{16}{c}{GARCH(1,0,1)} \\
    MGVBP & 0.043 & 0.230 &       & 0.737 &       & -2027.489 & -2017.633 & 25.560 & 5.124 & 1.637 & 0.330 & -609.837 & 4.772 & 1.577 & 1.423 & 0.382 \\
    MGVB  & 0.043 & 0.229 &       & 0.737 &       & -2027.493 & -2017.626 & 25.557 & 5.113 & 1.637 & 0.330 & -609.852 & 4.771 & 1.577 & 1.423 & 0.382 \\
    QBVI  & 0.043 & 0.229 &       & 0.737 &       & -2027.493 & -2017.626 & 25.557 & 5.114 & 1.637 & 0.330 & -609.852 & 4.771 & 1.577 & 1.423 & 0.382 \\
    MCMC  & 0.042 & 0.231 &       & 0.738 &       &       & -2017.713 & 25.586 & 5.189 & 1.637 & 0.330 & -609.678 & 4.779 & 1.574 & 1.422 & 0.381 \\
    ML    & 0.042 & 0.226 &       & 0.739 &       &       & -2017.593 & 25.555 & 5.073 & 1.637 & 0.331 & -610.093 & 4.762 & 1.581 & 1.424 & 0.384 \\
    \midrule
          & \multicolumn{16}{c}{GJR(1,1,1)} \\
    MGVBP & 0.043 & 0.108 & 0.294 & 0.722 &       & -2003.542 & -1990.907 & 27.125 & 7.039 & 1.606 & 0.310 & -603.476 & 5.464 & 1.556 & 1.393 & 0.346 \\
    MGVB  & 0.044 & 0.108 & 0.292 & 0.722 &       & -2003.550 & -1990.896 & 27.071 & 6.969 & 1.606 & 0.310 & -603.494 & 5.451 & 1.551 & 1.393 & 0.347 \\
    QBVI  & 0.044 & 0.108 & 0.292 & 0.722 &       & -2003.550 & -1990.897 & 27.073 & 6.971 & 1.606 & 0.310 & -603.492 & 5.451 & 1.551 & 1.393 & 0.347 \\
    MCMC  & 0.043 & 0.108 & 0.298 & 0.724 &       &       & -1991.095 & 27.339 & 7.318 & 1.606 & 0.309 & -603.116 & 5.504 & 1.576 & 1.391 & 0.344 \\
    ML    & 0.042 & 0.108 & 0.291 & 0.723 &       &       & -1990.853 & 27.048 & 6.940 & 1.606 & 0.311 & -603.851 & 5.441 & 1.548 & 1.394 & 0.349 \\
    \midrule
          & \multicolumn{16}{c}{GJR(1,1,2)} \\
    MGVBP & 0.045 & 0.116 & 0.323 & 0.649 & 0.054 & -2003.193 & -1990.888 & 27.657 & 7.517 & 1.606 & 0.311 & -604.242 & 5.623 & 1.604 & 1.396 & 0.348 \\
    MGVB  & 0.045 & 0.114 & 0.319 & 0.643 & 0.061 & -2003.247 & -1990.842 & 27.499 & 7.291 & 1.606 & 0.310 & -604.097 & 5.581 & 1.584 & 1.396 & 0.347 \\
    QBVI  & 0.045 & 0.114 & 0.319 & 0.643 & 0.060 & -2003.247 & -1990.843 & 27.501 & 7.294 & 1.606 & 0.310 & -604.098 & 5.582 & 1.585 & 1.396 & 0.347 \\
    MCMC  & 0.045 & 0.115 & 0.322 & 0.667 & 0.039 &       & -1990.928 & 27.690 & 7.615 & 1.606 & 0.311 & -604.171 & 5.631 & 1.611 & 1.396 & 0.348 \\
    ML    & 0.044 & 0.111 & 0.307 & 0.652 & 0.058 &       & -1990.807 & 27.236 & 6.979 & 1.606 & 0.311 & -603.993 & 5.500 & 1.555 & 1.395 & 0.348 \\
    \bottomrule
    \end{tabular}%
    }
  \label{tab:vol_est}
\end{table}%

\begin{table}[htbp]
  \centering
  \caption{Estimation results and performance metrics for EGARCH and FIGARCH models.}
  \scalebox{0.5}{
    \begin{tabular}{lccccccccccccccc}
          &       &       &       &       & \multicolumn{6}{c}{Train}                     & \multicolumn{5}{c}{Test} \\
          \cmidrule(lr){6-11} \cmidrule(lr){12-16}   
          & $\omega$ & $\alpha$ & $\gamma$ & $\beta_1$ & $\LB^\star$ & $p\br{y\vert \bmu^\star}$ & MSE\textsubscript{r} & MSE\textsubscript{rv} & QLIKE\textsubscript{r} & QLIKE\textsubscript{rv} & $p\br{y\vert \bmu^\star}$ & MSE\textsubscript{r} & MSE\textsubscript{rv} & QLIKE\textsubscript{r} & QLIKE\textsubscript{rv} \\
          \midrule
    \textit{\textbf{EGARCH(1,0,1)}} &       &       &       &       &       &       &       &       &       &       &       &       &       &       &  \\
    MGVBP & -0.003 & 0.413 &       & 0.929 & -2048.242 & -2032.756 & 26.713 & 5.156 & 1.655 & 0.337 & -610.894 & 4.740 & 1.667 & 1.428 & 0.381 \\
    MGVB  & -0.003 & 0.413 &       & 0.929 & -2048.242 & -2032.757 & 26.713 & 5.157 & 1.655 & 0.337 & -610.895 & 4.740 & 1.667 & 1.428 & 0.381 \\
    QBVI  & -0.003 & 0.416 &       & 0.929 & -2048.247 & -2032.779 & 26.709 & 5.154 & 1.655 & 0.337 & -610.911 & 4.742 & 1.666 & 1.428 & 0.381 \\
    MCMC  & -0.003 & 0.413 &       & 0.929 &       & -2032.752 & 26.711 & 5.155 & 1.655 & 0.337 & -610.878 & 4.739 & 1.667 & 1.428 & 0.381 \\
    ML    & -0.003 & 0.405 &       & 0.932 &       & -2032.723 & 26.732 & 5.169 & 1.655 & 0.337 & -610.840 & 4.730 & 1.672 & 1.427 & 0.381 \\
    \midrule
    \textit{\textbf{EGARCH(1,1,1)}} &       &       &       &       &       &       &       &       &       &       &       &       &       &       &  \\
    MGVBP & -0.014 & 0.350 & -0.172 & 0.930 & -2010.242 & -1989.666 & 26.039 & 4.413 & 1.604 & 0.303 & -600.116 & 4.972 & 1.356 & 1.377 & 0.327 \\
    MGVB  & -0.014 & 0.350 & -0.172 & 0.930 & -2010.242 & -1989.667 & 26.040 & 4.413 & 1.604 & 0.303 & -600.119 & 4.973 & 1.356 & 1.377 & 0.327 \\
    QBVI  & -0.015 & 0.364 & -0.175 & 0.926 & -2010.403 & -1989.914 & 26.081 & 4.454 & 1.604 & 0.304 & -600.654 & 5.022 & 1.363 & 1.379 & 0.328 \\
    MCMC  & -0.015 & 0.349 & -0.171 & 0.930 &       & -1989.665 & 26.051 & 4.419 & 1.604 & 0.303 & -600.112 & 4.970 & 1.356 & 1.377 & 0.327 \\
    ML    & -0.014 & 0.340 & -0.170 & 0.932 &       & -1989.610 & 26.011 & 4.386 & 1.604 & 0.303 & -599.682 & 4.935 & 1.352 & 1.375 & 0.327 \\
    \midrule
    \textit{\textbf{FIGARCH(0,d,1)}} & $\bar{\omega}$ & $\phi$ & $d$   & $\beta_1$ &       &       &       &       &       &       &       &       &       &       &  \\
    MGVBP & 0.099 &       & 0.644 & 0.415 & -2022.562 & -2015.130 & 25.533 & 5.540 & 1.634 & 0.326 & -609.607 & 4.789 & 1.574 & 1.422 & 0.382 \\
    MGVB  & 0.099 &       & 0.643 & 0.414 & -2022.562 & -2015.130 & 25.532 & 5.538 & 1.634 & 0.326 & -609.604 & 4.788 & 1.574 & 1.422 & 0.382 \\
    QBVI  & 0.099 &       & 0.643 & 0.414 & -2022.562 & -2015.130 & 25.532 & 5.539 & 1.634 & 0.326 & -609.607 & 4.789 & 1.574 & 1.422 & 0.382 \\
    MCMC  & 0.099 &       & 0.655 & 0.422 &       & -2015.186 & 25.537 & 5.566 & 1.634 & 0.327 & -609.886 & 4.800 & 1.574 & 1.423 & 0.383 \\
    ML    & 0.102 &       & 0.653 & 0.428 &       & -2015.094 & 25.582 & 5.570 & 1.634 & 0.326 & -609.345 & 4.780 & 1.574 & 1.420 & 0.381 \\
    \midrule
    \textit{\textbf{FIGARCH(1,d,1)}} &       &       &       &       &       &       &       &       &       &       &       &       &       &       &  \\
    MGVBP & 0.100 & 0.060 & 0.662 & 0.481 & -2022.666 & -2014.747 & 25.639 & 5.480 & 1.634 & 0.326 & -609.772 & 4.802 & 1.570 & 1.422 & 0.380 \\
    MGVB  & 0.100 & 0.060 & 0.661 & 0.480 & -2022.667 & -2014.746 & 25.637 & 5.478 & 1.634 & 0.326 & -609.761 & 4.802 & 1.570 & 1.422 & 0.380 \\
    QBVI  & 0.100 & 0.060 & 0.661 & 0.480 & -2022.667 & -2014.746 & 25.638 & 5.478 & 1.634 & 0.326 & -609.762 & 4.802 & 1.570 & 1.422 & 0.380 \\
    MCMC  & 0.100 & 0.059 & 0.668 & 0.482 &       & -2014.805 & 25.628 & 5.495 & 1.634 & 0.326 & -610.007 & 4.813 & 1.570 & 1.423 & 0.381 \\
    ML    & 0.100 & 0.064 & 0.653 & 0.479 &       & -2014.721 & 25.649 & 5.458 & 1.634 & 0.326 & -609.489 & 4.791 & 1.571 & 1.421 & 0.379 \\
    \bottomrule
    \end{tabular}%
    }
  \label{tab:vol_est_EGARCH}
\end{table}%

\begin{table}[htbp]
  \centering
  \caption{Additional results for the HAR model involving the $h$-function gradient estimation and the diagonal posterior form.}
  \scalebox{0.6}{
    \begin{tabular}{lcccccccccccc}
          &       &       &       &       &       & \multicolumn{4}{c}{Train}     & \multicolumn{3}{c}{Test} \\
\cmidrule(lr){7-10} \cmidrule(lr){11-13}   
& $\beta_0$ & $\beta_1$ & $\beta_2$ & $\beta_3$ & $\sigma_\varepsilon$ & $\LB^\star$ & $p\br{y\vert \bmu^\star}$ & MSE\textsubscript{rv} & $10^2\times$QLIKE\textsubscript{rv} & $p\br{y\vert \bmu^\star}$ & MSE\textsubscript{rv} & $10^2\times$QLIKE\textsubscript{rv} \\
\cmidrule{2-13}    MGVBP & 1.032 & 0.489 & 0.420 & -0.009 & 4.922 & -5080.191 & -5058.113 & 24.193 & 5.324 & -1220.154 & 18.861 & 5.712 \\
    MGVB  & 1.019 & 0.485 & 0.426 & -0.011 & 4.935 & -5082.123 & -5058.167 & 24.194 & 5.328 & -1220.550 & 18.879 & 5.720 \\
    QBVI  & 1.023 & 0.489 & 0.420 & -0.008 & 4.924 & -5080.679 & -5058.127 & 24.193 & 5.323 & -1220.206 & 18.862 & 5.714 \\
    MGVBP$^h\text{-func.}$ & 1.032 & 0.489 & 0.420 & -0.009 & 4.922 & -5080.192 & -5058.113 & 24.193 & 5.324 & -1220.148 & 18.861 & 5.712 \\
    MGVB$^h\text{-func.}$ & 1.019 & 0.485 & 0.426 & -0.011 & 4.934 & -5082.127 & -5058.166 & 24.194 & 5.328 & -1220.545 & 18.879 & 5.720 \\
    QBVI$^h\text{-func.}$ & 1.023 & 0.489 & 0.420 & -0.008 & 4.924 & -5080.652 & -5058.126 & 24.193 & 5.323 & -1220.200 & 18.862 & 5.714 \\
    MGVBP$^\text{diag.}$ & 1.042 & 0.487 & 0.423 & -0.010 & 4.925 & -5082.818 & -5058.111 & 24.193 & 5.327 & -1220.265 & 18.868 & 5.714 \\
    MGVB$^\text{diag.}$ & 1.020 & 0.485 & 0.430 & -0.015 & 4.944 & -5084.909 & -5058.230 & 24.195 & 5.331 & -1220.894 & 18.897 & 5.729 \\
    QBVI$^\text{diag.}$ & 1.036 & 0.485 & 0.423 & -0.007 & 4.931 & -5083.253 & -5058.143 & 24.193 & 5.326 & -1220.339 & 18.862 & 5.710 \\
    MCMC  & 1.075 & 0.488 & 0.421 & -0.012 & 4.922 &       & -5058.090 & 24.192 & 5.328 & -1220.130 & 18.858 & 5.706 \\
    ML    & 1.080 & 0.488 & 0.421 & -0.012 & 4.919 &       & -5058.086 & 24.192 & 5.328 & -1220.053 & 18.857 & 5.706 \\
    \bottomrule
    \end{tabular}%
    }
    \label{tab:vol_est_FIGARCH_HAR}
\end{table}%

\FloatBarrier
\subsection{Istanbul dataset: block-diagonal covariance}\label{app:istanbul}
In this section, we apply MGVBP under different assumptions for the structure of the variational covariance matrix. We use the Istanbul stock exchange dataset of \citep{akbilgic2014novel}, (details are provided in Appendix \ref{app:exp:hyperp}). To demonstrate the feasibility of the block-diagonal estimation under the mean-field framework outlined in Section \ref{subsec:mean_field}, we consider the following model for the daily returns of the Istanbul Stock Exchange National 100 index (ISE):
$$
\text{ISE}_t = \beta_0 + \beta_1 \text{SP}_t + \beta_2 \text{NIK}_t + \beta_3 \text{BOV}_t + \beta_4 \text{DAX}_t + \beta_5 \text{FTSE}_t + \beta_6 \text{EU}_t + \beta_7 \text{EM}_t + \varepsilon_t \text{,}
$$
with $\varepsilon_t \sim \N\br{0,\sigma_\varepsilon^2}$ (conditionally on the regressors at time $t$). The covariates respectively correspond to daily returns of the S\&P 500 index, the Japanese Nikkei index, the Brazilian Bovespa index, the German DAX index, the UK FTSE index, the MSCI European index, and of the MSCI emerging market index.  
We estimate the coefficients $\beta_0,\dots,\beta_7$ and the transformed parameter $\psi_\sigma = \log\br{\sigma}$, from which $\sigma$ is computed as $\sigma = \exp\br{\psi_\sigma} + \text{Var}\br{\psi_\sigma}/2$, with $\text{Var}\br{\psi_\sigma}$ taken from the variational posterior covariance matrix. 

We consider the following structures for the variational posterior: (i) full covariance matrix (\textit{Full}), (ii) diagonal covariance matrix (\textit{Diagonal}), (iii) block-diagonal structure with two blocks of sizes $8\times 8$ and $1\times 1$ (\textit{Block 1}) and, (iv) block diagonal structure with blocks of sizes $1\times 1$, $3 \times 3$, $2\times 2$, $2 \times 2$ and $1\times 1$ (\textit{Block 2}). Case (iii), models the covariance between the regression coefficients ($\beta$s) but neglects their covariance with the variance of the error $\sigma^2_\varepsilon$.
Case (iv) groups the indices traded in non-European stock exchanges in a $3\times 3$ block, and in the remaining $2 \times 2$ blocks, the indices referring to European exchanges and the two MSCI indexes. As for the earlier case, the covariances between the regression parameters and $\sigma^2_\varepsilon$ are ignored.
Note that the purpose of this application is to provide an example for Algorithm \ref{alg:mean_field} and for the discussion in Appendix \ref{subsec:mean_field}. To this end, structures (iii) and (iv) correspond to an intuitive and economically motivated grouping of the variables. Providing an effective predictive model supported by a solid econometric rationale is here out of scope.

Table \ref{tab:istanbul_mu} and \ref{tab:istanbul_cov} summarize the estimation results. Table \ref{tab:istanbul_mu} shows that the impact of the different structures of the covariance matrix is somewhat marginal in terms of the performance measures, with respect to each other and with respect to the ML estimates. As for the logistic regression example, in the most constrained cases (ii) and (iii), we observe that the estimates of certain posterior means slightly deviate from the others, indicating that the algorithm perhaps reaches a different maximum. Regarding the variational covariances reported in Table \ref{tab:istanbul_cov}, there is remarkable accordance between the covariance structures (i), (iii), and ML, while for the diagonal structure (ii) and block-diagonal structure (iv) the covariances are misaligned with the ML and full-diagonal case, further suggesting the convergence of the algorithms at different maxima of the LB.

\begin{figure}
    \centering
    \includegraphics[trim={0cm 0.5cm 0cm 0cm},clip,width = 0.7\textwidth]{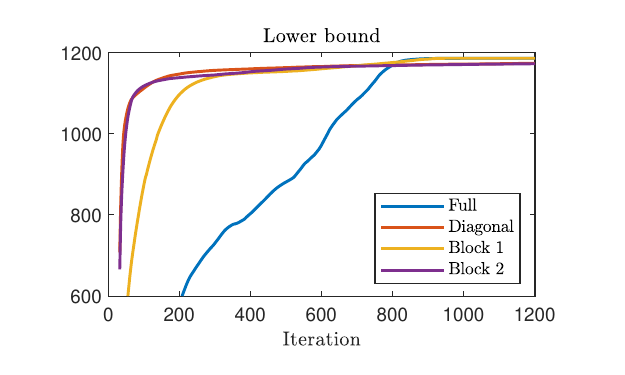}
    \caption{Lower bound optimization for the Istanbul data under the different posterior variational covariance matrix specifications.}
    \label{fig:istanbul_elbo}
\end{figure}

From a theoretical perspective, if $\S$ is the covariance matrix of the joint  Gaussian distribution of the variates (case (i)), estimates of the block-diagonal entries (or main diagonal) should match the corresponding elements in $\S$. However, the elements in the sub-matrices, e.g., in cases (ii) and (iv), deviate from those $\S$, case (i). 
Indeed, the results refer to independent optimizations of alternative models (over the same dataset) that are not granted to converge at the same LB maximum (and thus variational Gaussian). Across the covariance structures (i) to (iv) the optimal variational parameters correspond to different multivariate distributions, that independently maximize the lower bound, and that are not constrained to be related to each other. This is indeed confirmed by the differences in the maximized Lower bound $\LB\br{\bz^\star}$ in Table \ref{tab:istanbul_mu}, and in the different levels at which the curves in Figure \ref{fig:istanbul_elbo} are observed to converge. 
In this light, the ML estimates' variances in Table \ref{tab:istanbul_cov_2} can be compared to those of case (i), but are misleading for the other cases, as the covariance matrix of the asymptotic (Gaussian) distribution of the ML estimator is implicitly a full matrix.

\begin{table}
    \centering
        \caption{Posterior means, ML estimates, and performance measures on the train and test set.}
    \scalebox{0.7}{
\begin{tabular}{lSSSSSSSSS}
& \multicolumn {1}{c}{$\beta_0$} & \multicolumn {1}{c}{$\beta_1$} & \multicolumn {1}{c}{$\beta_2$} & \multicolumn {1}{c}{$\beta_3$} & \multicolumn {1}{c}{$\beta_4$} & \multicolumn {1}{c}{$\beta_5$} & \multicolumn {1}{c}{$\beta_6$} & \multicolumn {1}{c}{$\beta_7$} & \multicolumn {1}{c}{$\sigma_\varepsilon$} \\
\midrule
    Full (case i) & 0.001 & 0.098 & 0.079 & -0.271 & -0.167 & -0.354 & 1.164 & 0.944 & 0.014 \\
    Diagonal (case ii) & 0.001 & 0.074 & 0.092 & -0.239 & 0.093 & -0.015 & 0.555 & 0.935 & 0.014 \\
    Block 1 (case iii) & 0.001 & 0.098 & 0.079 & -0.272 & -0.167 & -0.353 & 1.164 & 0.943 & 0.014 \\
    Block 2 (case iv) & 0.001 & 0.067 & 0.115 & -0.218 & 0.201 & 0.162 & 0.293 & 0.871 & 0.014 \\
    ML    & 0.001 & 0.099 & 0.078 & -0.273 & -0.174 & -0.363 & 1.179 & 0.946 & 0.014 \\
\bottomrule
\end{tabular}
}
\\
    \scalebox{0.7}{
\begin{tabular}{lSSSSS}
      & \multicolumn {3}{c}{Train} & \multicolumn {2}{c}{Test} \\
\cmidrule(lr){2-4}  \cmidrule(lr){5-6}     & \multicolumn {1}{c}{$\LB^\star$} & \multicolumn {1}{c}{$p\br{y\vert \bmu^\star}$}    & \multicolumn {1}{c}{$10^2 \times$MSE }   & \multicolumn {1}{c}{$p\br{y\vert \bmu^\star}$}    & \multicolumn {1}{c}{$10^2 \times$MSE } \\
\midrule
    Full  & 1186.082 & 1223.70 & 19.41 & 316.84 & 4.15 \\
    Diagonal & 1173.662 & 1214.83 & 20.23 & 316.71 & 4.12 \\
    Block 1 & 1186.087 & 1223.70 & 19.41 & 316.86 & 4.15 \\
    Block 2 & 1172.580 & 1212.28 & 20.48 & 316.72 & 4.10 \\
    ML    &       & 1223.73 & 19.41 & 316.87 & 4.15 \\
\bottomrule
\end{tabular}
}
    \label{tab:istanbul_mu}
\end{table}

\begin{table}
    \centering
        \caption{Top panel: posterior covariance matrix under the full specification (case i), and covariances of the ML estimates. Bottom panel: posterior covariance matrices under the Block 1 and Block 2 structures. All the entries are multiplied by $10^4$.}
\scalebox{0.7}{   
\begin{tabular}{ccSSSSSSSSS}
&       & \multicolumn {9}{c}{Full} \\
\cmidrule{3-11}      &       & \multicolumn {1}{c}{$\beta_1$} & \multicolumn {1}{c}{$\beta_2$} & \multicolumn {1}{c}{$\beta_3$} & \multicolumn {1}{c}{$\beta_4$} & \multicolumn {1}{c}{$\beta_5$} & \multicolumn {1}{c}{$\beta_6$} & \multicolumn {1}{c}{$\beta_7$} &\multicolumn {1}{c}{$\beta_8$} & \multicolumn {1}{c}{$\sigma_\varepsilon$} \\
\midrule
\multirow  {8}[2]{*}{\rotatebox[origin=c]{90}{ML}} 
         & $\beta_0$ &       & 0.001 & 0.002 & -0.001 & -0.002 & -0.001 & 0.006 & -0.008 & 0.000 \\
          & $\beta_1$ & 0.001 &       & 0.008 & -2.947 & -1.915 & -0.767 & -0.094 & 1.520 & -0.001 \\
          & $\beta_2$ & 0.003 & -0.032 &       & 1.094 & 0.442 & 0.768 & -0.640 & -4.141 & 0.012 \\
          & $\beta_3$ & -0.001 & -2.992 & 1.094 &       & 1.018 & 0.683 & -1.213 & -4.710 & 0.017 \\
          & $\beta_4$ & -0.002 & -1.906 & 0.297 & 0.961 &       & 3.742 & -20.149 & -0.748 & 0.028 \\
          & $\beta_5$ & -0.002 & -0.729 & 0.679 & 0.622 & 3.836 &       & -28.773 & -1.594 & 0.052 \\
          & $\beta_6$ & 0.007 & -0.170 & -0.393 & -1.065 & -20.477 & -28.813 &       & -2.984 & -0.142 \\
          & $\beta_7$ & -0.009 & 1.605 & -4.148 & -4.700 & -0.563 & -1.386 & -3.324 &       & 0.030 \\
\midrule
      &       &       &       &       &       &       &       &       &       &  \\
&       & \multicolumn {9}{c}{Block 1} \\
\cmidrule{3-11}      &       & \multicolumn {1}{c}{$\beta_1$} & \multicolumn {1}{c}{$\beta_2$} & \multicolumn {1}{c}{$\beta_3$} & \multicolumn {1}{c}{$\beta_4$} & \multicolumn {1}{c}{$\beta_5$} & \multicolumn {1}{c}{$\beta_6$} & \multicolumn {1}{c}{$\beta_7$} &\multicolumn {1}{c}{$\beta_8$} & \multicolumn {1}{c}{$\sigma_\varepsilon$} \\
\midrule
\multirow  {8}[2]{*}{\rotatebox[origin=c]{90}{Block 2}} & $\beta_0$ &       & 0.001 & 0.002 & -0.001 & -0.003 & -0.002 & 0.008 & -0.009 &  \\
          & $\beta_1$ &       &       & -0.064 & -2.966 & -2.005 & -0.728 & -0.049 & 1.665 &  \\
          & $\beta_2$ &       & 0.322 &       & 1.079 & 0.296 & 0.690 & -0.382 & -4.075 &  \\
          & $\beta_3$ &       & -3.327 & -0.525 &       & 0.950 & 0.512 & -1.010 & -4.558 &  \\
          & $\beta_4$ &       &       &       &       &       & 3.841 & -20.608 & -0.584 &  \\
          & $\beta_5$ &       &       &       &       & -8.708 &       & -28.509 & -1.107 &  \\
          & $\beta_6$ &       &       &       &       &       &       &       & -3.584 &  \\
          & $\beta_7$ &       &       &       &       &       &       & -4.452 &       &  \\
\bottomrule
      &       &       &       &       &       &       &       &       &       &  \\
\end{tabular}
}
    \label{tab:istanbul_cov}
\end{table}

\begin{table} 
\centering
\caption{Standard deviations of the posterior marginals for the different cases, along with ML standard errors. Entries are multiplied by $10^2$.}
\scalebox{0.7}{
\begin{tabular}{lSSSSSSSSSS}
      & \multicolumn {1}{c}{$\beta_1$} & \multicolumn {1}{c}{$\beta_2$} & \multicolumn {1}{c}{$\beta_3$} & \multicolumn {1}{c}{$\beta_4$} & \multicolumn {1}{c}{$\beta_5$} & \multicolumn {1}{c}{$\beta_6$} & \multicolumn {1}{c}{$\beta_7$} &\multicolumn {1}{c}{$\beta_8$} & \multicolumn {1}{c}{$\sigma$} \\
\midrule
    Full  & 0.069 & 7.472 & 5.641 & 7.322 & 12.915 & 16.663 & 23.056 & 12.395 & 3.463 \\
    Diagonal & 0.073 & 4.662 & 4.056 & 4.477 & 4.343 & 5.587 & 4.733 & 5.976 & 3.599 \\
    Block 1 & 0.068 & 7.470 & 5.629 & 7.274 & 13.072 & 16.534 & 23.092 & 12.257 & 3.421 \\
    Block 2 & 0.071 & 6.530 & 4.451 & 6.611 & 9.333 & 10.620 & 7.001 & 8.724 & 3.367 \\
    ML    & 0.068 & 7.489 & 5.631 & 7.322 & 12.997 & 16.636 & 23.133 & 12.363 &  \\
    \bottomrule
\end{tabular}%
}
    \label{tab:istanbul_cov_2}
\end{table}

\FloatBarrier

\FloatBarrier
\subsection{Datasets and hyperparameters}\label{app:exp:hyperp}

\begin{table}[ht]
    \centering
        \caption{Details on the datasets and corresponding models (upper panel), and hyperparameters used in the experiments (lower panel).}
    \scalebox{0.7}{
    \begin{tabular}{llccccl}
    \multicolumn {1}{p{3.72em}}{\centering Dataset} & \multicolumn {1}{p{7.835em}}{\centering Model} & \multicolumn {1}{p{5em}}{\centering Number of variational parameters} & \multicolumn {1}{p{5em}}{\centering Number of samples} & \multicolumn {1}{p{5em}}{\centering Samples in train set} & \multicolumn {1}{p{5em}}{\centering Samples in test set} & \multicolumn {1}{p{10.72em}}{\centering Period} \\
    \midrule
    Labour & Logistic regression & 72     & 753   & 564 (75\%) & 189 (25\%) &  \\
    S\&P 500 & ARCH  & 6     & 2123  & 1689 (80\%) & 425 & 3-Jan-2014 / 28-Jun-2022 \\
          & GARCH(1,0,1) & 12     & 2123  & 1689  & 425 & 3-Jan-2014 / 28-Jun-2022 \\
          & GJR(1,1,1) & 20     & 2123  & 1689  & 425  & 3-Jan-2014 / 28-Jun-2022 \\
          & GJR(1,1,2) & 30     & 2123  & 1689  & 425  &  3-Jan-2014 / 28-Jun-2022\\
          & EGARCH(1,0,1) & 12     & 2123  & 1689 & 425  & 3-Jan-2014 / 28-Jun-2022 \\
          & EGARCH(1,1,1) & 20     & 2123  & 1689  & 425  & 3-Jan-2014 / 28-Jun-2022 \\
          & EGARCH(1,1,2) & 30    & 2123  & 1689  & 425  & 3-Jan-2014 / 28-Jun-2022 \\
          & FIGARCH(0,1,1) & 12     & 2123  & 1689  & 425  & 3-Jan-2014 / 28-Jun-2022 \\
          & FIGARCH(1,1,2) & 20     & 2123  & 1689  & 425  & 3-Jan-2014 / 28-Jun-2022 \\
          & HAR (Linear regr.) & 30   & 2102  & 1681  & 421 & 4-Feb-2014 / 28-Jun-2022 \\          
    Istanbul & Linear regression & 90   & 536   & 428 (80\%) & 108  & 5-Jan-2009 / 22-Feb-2022 \\
    LOB & Neural network & 54990 & 557297 & 256461(65\%) & 150418 & 1-Jun-2010 / 14-Jun-2010\\
    Synthetic & Non-diff. model & 420 & 5000 & 4000 (80\%) & 1000 \\
    \bottomrule
    \end{tabular}
}
\vspace{0.5cm}

\scalebox{0.6}{
\begin{tabular}{lccccccccccccc}
      & \multicolumn {9}{c}{MGVBP optimizer }                                        & \multicolumn {2}{c}{Initial values} & \multicolumn {2}{c}{Prior} \\
\cmidrule(lr){2-10} \cmidrule(lr){11-12} \cmidrule(lr){13-14}
\multicolumn {1}{c}{Experiment} & $\beta$ & Grad. clip & Grad. clip init. & $\omega$ & $w$   & $t_\text{max}$ & $t'$  & $P$   & $S$   & $\bmu_1$ & $\S_1$ & $\bmu_0$ & $\S_0$ \\
\midrule
Labour data    & 0.01  & 3000  & 1000  & 0.4   & 30    & 1200  & 1000      & 500  & 75    & $\sim\N\br{0,\S_1}$        & 0.05  & 0     & 5 \\
ARCH-GARCH-GJR  & 0.01  & 1000  & 1000  & 0.4   & 30    & 1200  & 1000      & 500  & 150   & ML                         & 0.05  & 0     & 5 \\
EGARCH          & 0.01  & 1000  & 1000  & 0.4   & 30    & 3000  & 2500      & 500  & 150   & ML                         & 0.05  & 0     & 5 \\
FIGARCH         & 0.01  & 1000  & 1000  & 0.4   & 30    & 1200  & 1000      & 500  & 150   & ML                         & 0.05  & 0     & 5 \\
HAR             & 0.001 & 50000 & 1000   & 0.4   & 30    & 1000  & 750      & 100  & 150    & ML                         & 0.01  & 0     & 5 \\
Istanbul data   & 0.07  & 50000 & 500   & 0.4   & 30    & 1200  & 1000      & 500  & 100   & $\sim\N\br{0,\S_1}$        & 0.01  & 0     & 5 \\
LOB data   & 0.1  & 5000 & 100   & 0.4   & 30    & 4000  & 3000      & 500  & 20   & ADAM        & 0.01  & 0     & 5 \\
Non-diff. model &0.1 &150 &1000 &0.4 &30 &5000 & 5000 &500 &15 & $\sim\N\br{0,\S_1}$ & 0.01 &0&1 \\
\bottomrule
\end{tabular}
}
    \label{tab:hyperparameters}
\end{table}

Table \ref{tab:hyperparameters} summarizes some information about the datasets and the setup used across the experiments. For the experiments on the Labour and S\&P 500 datasets, the same set of hyperparameters applies to MGVBP, MGVB (and QBVI).
While the Labour\footnote{Publicly available at \url{key2stats.com/data-set/view/140}. See \citep{mroz1984sensitivity} for details. The data is also adopted in other VI applications, e.g., \citep{tran2021practical,magris2022quasi}.} and Istanbul\footnote{Publicly available at the UCI repository, \url{archive.ics.uci.edu/ml/datasets/istanbul+stock+exchange}. See \citep{akbilgic2014novel} for details.} datasets are readily available, the S\&P 500 dataset is extracted from the Oxford-Man Institute realized volatility library\footnote{\url{realized.oxford-man.ox.ac.uk}.}. We use daily close-to-close demeaned returns for the GARCH-family models and 5-minute sub-sampled daily measures of realized volatilities (further annualized) for the HAR model. For information on the publicly available limit-order book (LOB) dataset, see \citep{ntakarislob2018}. MGVBP and all optimizers appearing in Table \ref{tab:tabl} are initialized after 30 ADAM iterations for the neural network experiments. VOGN is implemented on the LOB data following the scheme presented in \citep{magris2022bayesian}. The synthetic data used for the non-differentiable model is discussed within Section \ref{sec:experiments}. 

\subsubsection*{Starting values}
As a robustness check, in Table \ref{tab:StartValLabour} and Table \ref{tab:StartValGARCH} we provide summary statistics of the estimation results and performance metrics for two representative models. With both a random and constant initialization of $\bth_0$, we see a tight consistency in the results, underlining the robustness of the estimation routine with respect to the choice of the starting values.

\begin{table}[htbp]
  \centering
  \caption{Estimation results for the Logistic regression model (Labour dataset) under different starting values. Statistics are computed across 200 replications. Upper table: random initialization of $\bth_0$ from a multivariate Gaussian with diagonal covariance matrix. Lower table: initialization of $\bth_0$ as a vector of constants.}
  \scalebox{0.60}{
    \begin{tabular}{cccccccccccc}
    \multicolumn{2}{c}{$\bth_0 \sim \mathcal{N}(\bmu,\sigma^2I)$} & \multicolumn{5}{c}{Mean}              & \multicolumn{5}{c}{Std. Dev. $(\times 100)$} \\
   \cmidrule(lr){1-2} \cmidrule(lr){3-7} \cmidrule(lr){8-12}
    $\bmu$ & $\sigma^2$ & \multicolumn{1}{c}{$\LB\br{\bz^\star}$} & \multicolumn{1}{c}{Accuracy} & \multicolumn{1}{c}{Precision} & \multicolumn{1}{c}{Recall} & \multicolumn{1}{c}{F1} & \multicolumn{1}{c}{$\LB\br{\bz^\star}$} & \multicolumn{1}{c}{Accuracy} & \multicolumn{1}{c}{Precision} & \multicolumn{1}{c}{Recall} & \multicolumn{1}{c}{F1} \\
    \midrule
    0     & 0.1   & -356.640 & 71.179 & 71.108 & 70.216 & 70.659 & 0.091 & 0.090 & 0.103 & 0.084 & 0.093 \\
    0     & 1     & -356.640 & 71.152 & 71.076 & 70.194 & 70.632 & 0.081 & 0.083 & 0.096 & 0.076 & 0.086 \\
    0     & 5     & -356.640 & 71.117 & 71.037 & 70.160 & 70.595 & 0.117 & 0.079 & 0.087 & 0.077 & 0.082 \\
    0     & 10    & -356.641 & 71.126 & 71.047 & 70.168 & 70.605 & 0.065 & 0.065 & 0.074 & 0.060 & 0.067 \\
    5     & 0.1   & -356.640 & 71.117 & 71.035 & 70.161 & 70.596 & 0.111 & 0.055 & 0.063 & 0.049 & 0.056 \\
    5     & 1     & -356.640 & 71.126 & 71.047 & 70.168 & 70.605 & 0.117 & 0.087 & 0.096 & 0.084 & 0.089 \\
    5     & 5     & -356.640 & 71.137 & 71.061 & 70.175 & 70.616 & 0.081 & 0.095 & 0.105 & 0.092 & 0.098 \\
    5     & 10    & -356.640 & 71.183 & 71.111 & 70.222 & 70.664 & 0.072 & 0.091 & 0.105 & 0.082 & 0.093 \\
    -5    & 0.1   & -356.640 & 71.117 & 71.035 & 70.161 & 70.596 & 0.113 & 0.055 & 0.063 & 0.049 & 0.056 \\
    -5    & 1     & -356.640 & 71.126 & 71.045 & 70.170 & 70.605 & 0.090 & 0.065 & 0.075 & 0.059 & 0.067 \\
    -5    & 5     & -356.640 & 71.152 & 71.078 & 70.192 & 70.632 & 0.098 & 0.101 & 0.113 & 0.097 & 0.104 \\
    -5    & 10    & -356.640 & 71.146 & 71.068 & 70.188 & 70.625 & 0.086 & 0.080 & 0.092 & 0.073 & 0.082 \\
    \midrule
          &       &       &       &       &       &       &       &       &       &       &  \\
         \multicolumn{2}{c}{$\bth_0 = c\bm{1}_d$}       & \multicolumn{5}{c}{Mean}              & \multicolumn{5}{c}{Std. Dev. $(\times 100)$} \\
 \cmidrule(lr){1-2} \cmidrule(lr){3-7} \cmidrule(lr){8-12}
 \multicolumn{2}{c}{$c$ }       & \multicolumn{1}{c}{$\LB\br{\bz^\star}$} & \multicolumn{1}{c}{Accuracy} & \multicolumn{1}{c}{Precision} & \multicolumn{1}{c}{Recall} & \multicolumn{1}{c}{F1} & \multicolumn{1}{c}{$\LB\br{\bz^\star}$} & \multicolumn{1}{c}{Accuracy} & \multicolumn{1}{c}{Precision} & \multicolumn{1}{c}{Recall} & \multicolumn{1}{c}{F1} \\
    \midrule
     \multicolumn{2}{c}{0     }       & -356.640 & 71.165 & 71.092 & 70.203 & 70.644 & 0.064 & 0.088 & 0.100 & 0.082 & 0.090 \\
     \multicolumn{2}{c}{1     }       & -356.640 & 71.146 & 71.068 & 70.188 & 70.625 & 0.080 & 0.080 & 0.092 & 0.073 & 0.082 \\
    \multicolumn{2}{c}{ -1    }       & -356.640 & 71.144 & 71.067 & 70.184 & 70.623 & 0.126 & 0.098 & 0.108 & 0.093 & 0.100 \\
    \multicolumn{2}{c}{ 5     }       & -356.640 & 71.137 & 71.059 & 70.177 & 70.616 & 0.080 & 0.095 & 0.105 & 0.091 & 0.098 \\
     \multicolumn{2}{c}{-5    }       & -356.640 & 71.144 & 71.066 & 70.186 & 70.623 & 0.077 & 0.079 & 0.091 & 0.071 & 0.081 \\
     \multicolumn{2}{c}{20    }       & -356.641 & 71.135 & 71.057 & 70.176 & 70.614 & 0.112 & 0.073 & 0.083 & 0.067 & 0.075 \\
     \multicolumn{2}{c}{-20   }       & -356.641 & 71.179 & 71.108 & 70.216 & 70.659 & 0.114 & 0.107 & 0.119 & 0.102 & 0.110 \\
    \bottomrule
    \end{tabular}%
    }
  \label{tab:StartValLabour}%
\end{table}%

\begin{table}[htbp]
  \centering
  \caption{Estimation results for the GARCH(1,0,1) model under different starting values. Statistics are computed across 200 replications. Upper table: random initialization of $\bth_0$ from a multivariate Gaussian with diagonal covariance matrix. Lower table: initialization of $\bth_0$ as a vector of constants.}
  \scalebox{0.60}{
    \begin{tabular}{cccccccc}
    \multicolumn{2}{c}{$\bth_0 \sim \mathcal{N}(\bmu,\sigma^2I)$} & \multicolumn{3}{c}{Mean} & \multicolumn{3}{c}{Std. Dev. $(\times 100)$} \\
    \cmidrule(lr){1-2} \cmidrule(lr){3-5} \cmidrule(lr){6-8}
    $\bmu$ & $\sigma^2$ & \multicolumn{1}{l}{$\LB\br{\bz^\star}$} & $p\br{\by\vert \bth^\star}$ & MSE &$\LB\br{\bz^\star}$  & $p\br{\by\vert \bth^\star}$ & MSE \\
    \midrule
    0     & 0.1   & -2012.404 & 2002.560 & 25.690 & 0.150 & 0.243 & 0.122 \\
    0     & 1     & -2012.404 & 2002.560 & 25.690 & 0.148 & 0.252 & 0.129 \\
    0     & 5     & -2012.404 & 2002.561 & 25.691 & 0.144 & 0.259 & 0.112 \\
    0     & 10    & -2012.412 & 2002.564 & 25.689 & 3.046 & 0.744 & 1.245 \\
    5     & 0.1   & -2012.404 & 2002.562 & 25.692 & 0.156 & 0.345 & 0.160 \\
    5     & 1     & -2012.404 & 2002.562 & 25.691 & 0.156 & 0.348 & 0.188 \\
    5     & 5     & -2012.404 & 2002.563 & 25.692 & 0.146 & 0.603 & 0.280 \\
    5     & 10    & -2012.405 & 2002.562 & 25.691 & 0.179 & 0.557 & 0.315 \\
    -5    & 0.1   & -2012.404 & 2002.563 & 25.691 & 0.136 & 0.292 & 0.129 \\
    -5    & 1     & -2012.404 & 2002.563 & 25.691 & 0.134 & 0.646 & 0.240 \\
    -5    & 5     & -2012.404 & 2002.563 & 25.691 & 0.147 & 0.513 & 0.092 \\
    -5    & 10    & -2012.405 & 2002.565 & 25.693 & 0.216 & 1.042 & 0.450 \\
    \midrule
          &       &       &       &       &       &       &  \\
          \multicolumn{2}{c}{$\bth_0 = c\bm{1}_d$}& \multicolumn{3}{c}{Mean} & \multicolumn{3}{c}{Std. Dev. $(\times 100)$} \\
 \cmidrule(lr){1-2} \cmidrule(lr){3-5} \cmidrule(lr){6-8}
\multicolumn{2}{c}{$c$}      & $\LB\br{\bz^\star}$  & $p\br{\by\vert \bth^\star}$ & MSE & $\LB\br{\bz^\star}$  & $p\br{\by\vert \bth^\star}$ & MSE \\
    \midrule
   \multicolumn{2}{c}{ 0 }          & -2012.403 & 2002.560 & 25.691 & 0.234 & 0.212 & 0.107 \\
    \multicolumn{2}{c}{1    }       & -2012.403 & 2002.560 & 25.691 & 0.233 & 0.191 & 0.099 \\
    \multicolumn{2}{c}{-1    }       & -2012.403 & 2002.560 & 25.691 & 0.233 & 0.177 & 0.089 \\
    \multicolumn{2}{c}{5     }       & -2012.403 & 2002.562 & 25.692 & 0.238 & 0.401 & 0.188 \\
    \multicolumn{2}{c}{-5    }       & -2012.404 & 2002.562 & 25.691 & 0.229 & 0.302 & 0.161 \\
    \multicolumn{2}{c}{20    }       & -2012.405 & 2002.563 & 25.691 & 0.254 & 0.822 & 0.472 \\
    \multicolumn{2}{c}{-20   }       & -2012.402 & 2002.563 & 25.691 & 0.206 & 0.180 & 0.115 \\
    \bottomrule
    \end{tabular}%
    }
  \label{tab:StartValGARCH}%
\end{table}%

\subsubsection*{Runtime}
Table \ref{tab:runtime} compares the running times of the adopted Variational Bayes algorithms for two representative models. As expected, MGVBP and MGVB show a rather aligned performance (the factor 2 in \eqref{eq:update_iS_martin} compared to \eqref{eq:tran_updates} is indeed irrelevant in terms of running time). On the other hand, QBVI relies on update rules that are completely unrelated and different from those of MGVBP and MGVB; as such, the corresponding running times are provided as a reference since they are not directly comparable. Irrespective of the adopted optimizers, the major factor deriving the runtime appears to be the complexity of the loglikelihood (with little surprise the iterative nature of the expensive computations in the GARCH likelihood severely affects the runtime) the $\log p$ estimator variant (eq. \eqref{eq:estim_gaussian}) is consistently outperforming the classical approach based on the $h$-function estimator (eq. \eqref{eq:estim_general}).

\begin{table}[htbp]
  \centering
    \caption{Running time for the logistic regression (Labour dataset). Statistics in seconds per 1000 iterations averaged over 200 replications ($N_s = 75$).}
    \scalebox{0.7}{
    \begin{tabular}{cclccccccrrrrrr}
          &       &       & \multicolumn{6}{c}{Logistic regression}       & \multicolumn{3}{c}{GARCH(1,0,1)} & \multicolumn{3}{c}{GJR(1,1,1)} \\
\cmidrule(lr){4-9}      \cmidrule(lr){10-12} \cmidrule(lr){13-15}    &       &       & Mean  & Median & Std.  & Mean  & Median & Std.   & Mean  & Median & Std.  & Mean  & Median & Std. \\
\cmidrule(lr){4-6}  \cmidrule(lr){7-9}   \cmidrule(lr){10-12}    \cmidrule(lr){13-15}        
    \multicolumn{2}{c}{\multirow{4}[1]{*}{\rotatebox[origin=c]{90}{\parbox[c]{2cm}{\centering $h$-func. estimator}}}} &       & \multicolumn{3}{l}{\textit{Full cov.}} & \multicolumn{3}{l}{\textit{Diagonal cov.}} & \multicolumn{3}{l}{\textit{Full cov.}} & \multicolumn{3}{l}{\textit{Full cov.}} \\
    \multicolumn{2}{c}{} & MGVB & 5.411 & 5.387 & 1.137 & 5.375 & 5.346 & 0.181 & 32.76 & 33.05 & 0.78  & 30.73 & 30.45 & 2.44\\ 
    \multicolumn{2}{c}{} & MGVBP & 5.375 & 5.368 & 1.483 & 5.321 & 5.328 & 0.127 & 33.02 & 32.90 & 0.45  & 30.91 & 32.07 & 2.30 \\
    \multicolumn{2}{c}{} & QBVI  & 5.340 & 5.305 & 0.884 & 5.425 & 5.408 & 0.254 & 32.70 & 32.67 & 0.66  & 30.79 & 31.03 & 2.53 \\
\cmidrule{3-15}          
    \multicolumn{2}{c}{\multirow{4}[1]{*}{\rotatebox[origin=c]{90}{\parbox[c]{2cm}{\centering $\log p$ estimator}}}} &       & \multicolumn{3}{l}{\textit{Full cov.}} & \multicolumn{3}{l}{\textit{Diagonal cov.}} & \multicolumn{3}{l}{\textit{Full cov.}} & \multicolumn{3}{l}{\textit{Full cov.}} \\
    \multicolumn{2}{c}{} & MGVB & 5.883 & 5.877 & 0.121 & 5.791 & 5.733 & 0.419 & 30.94 & 31.18 & 1.52  & 29.18 & 30.18 & 1.89 \\
    \multicolumn{2}{c}{} & MGVBP & 5.929 & 5.904 & 0.184 & 5.666 & 5.656 & 0.743 & 31.47 & 31.69 & 0.82  & 29.21 & 29.79 & 1.72 \\
    \multicolumn{2}{c}{} & QBVI  & 5.879 & 5.858 & 0.119 & 5.543 & 5.551 & 0.353 & 31.14 & 31.39 & 0.94  & 28.86 & 29.98 & 2.39 \\
\cmidrule{3-15}    
\end{tabular}%
}
  \label{tab:runtime}%
\end{table}%

Table \ref{tab:runtime_npar} reports the runtime for a linear regression experiment (simulated data, $N_s = 100$, averaged across 50 replications) per iteration with a growing number of variational parameters. It can be seen that MGVBP is slightly but consistently faster than MGVB. As discussed in Section \ref{subsec:computational}, despite the simple natural gradient form the MGVBP has, the actual computational effort is a retraction, which requires inverting the variational covariance matrix at each iteration. Nevertheless, the experiments we performed show a steeper LB optimization and an improved optimum.

\begin{table}[htbp]
  \centering
  \caption{Running times (seconds per iteration) for a linear regression problem of increasing complexity.}
  \scalebox{0.7}{  
    \begin{tabular}{lccccccccc}
    Number of parameters & 6     & 12    & 30    & 110   & 240   & 650   & 2550  & 10100 & 22650 \\
    \midrule
    MGVB  & 0.0365 & 0.0174 & 0.0194 & 0.0247 & 0.0323 & 0.0766 & 0.2112 & 0.6997 & 1.5176 \\
    MGVBP & 0.0355 & 0.0169 & 0.0185 & 0.0237 & 0.0308 & 0.0760 & 0.2044 & 0.6795 & 1.5037 \\
    \midrule
    Difference  & 2.8\% & 2.8\% & 4.6\% & 4.0\% & 4.7\% & 0.8\% & 3.2\% & 2.9\% & 0.9\% \\
    \bottomrule
    \end{tabular}%
    }
  \label{tab:runtime_npar}%
\end{table}%

\section{Proofs}\label{app:proofs}

\subsection{Preliminaries: the Gaussian FIM}\label{app:gaussian_fim}

From \citep{barfoot2020multivariate} the natural gradients for a $d$-variate Gaussian distribution with mean $\bmu$ and covariance matrix $\S$, in the parametrizations $\bz$ and $\bzp$ are respectively given by:
\begin{align} 
            \mathcal{I}^{-1}_{\bz} = 
    \begin{bmatrix}
        \S &\bm{0}\\
        \bm{0} & 2\br{\S \otimes \S}
    \end{bmatrix}\text{,}\qquad    
            \mathcal{I}^{-1}_{\bzp}= 
    \begin{bmatrix}
        \S &\bm{0}\\
        \bm{0} & 2\br{\S \otimes \S}^{-1}
    \end{bmatrix}\text{.}\label{eq:barfoot}
\end{align}

\subsection{Preliminaries: a useful relation}
For a generic SPD matrix $P$ and a function $f$ of $P$ that can be as well reparametrized in terms of $P^{-1}$:
\begin{equation}\label{eq:riemann_grad_M}
 P \nabla_P f P = -\nabla_{P^{-1}} f  \text{,}
\end{equation}
which can be shown to hold according to matrix calculus with Jacobian transformations of the gradient $\nabla_P f$.

\subsection{Proof of Proposition \ref{prop:natgrad}} \label{app:proof_prop}
From the natural gradient definition:
\begin{align*}
\natgrad_{\bzp}\LB^i \stackrel{\text{def.}}{=}  \mathcal{I}^{-1}_{\bzp} \nabla_{\iS}\LB^i \stackrel{\eqref{eq:barfoot}}{=}  
    \begin{bmatrix}
        \S \nabla_\bmu \LB^i\\
        2\br{\S \otimes \S}^{-1} \nabla_\iS \LB^i
    \end{bmatrix}\text{.}
\end{align*}
removing the vectorization from the term related to $\iS$:
\begin{equation}
    \natgrad_{\iS}\LB^i = 2\iS \nabla_{\iS} \LB^i \iS \stackrel{\eqref{eq:riemann_grad_M}}{=} -2\nabla_\S \LB^c\text{.}\label{eq:equiv}
\end{equation}

\subsection{Derivation of the MGVBP update}\label{app:justification}
For the precision matrix, the Riemann gradient from the manifold $\mathcal{M}$ is $\iS \nabla_{\iS} \iS$. Therefore, the following equivalent representations hold:
$$
\bar{\nabla}_{\iS}\LB \stackrel{\eqref{eq:riemann_grad_M}}{=}  \iS \nabla_{\iS} \iS \stackrel{\eqref{eq:riemann_grad_M}}{=} -\nabla_\S \LB \stackrel{\eqref{eq:equiv}}{=} \frac{1}{2}\natgrad_{\iS}\LB\text{.}
$$
Note that the last equality is not general; it holds because of the particular $\br{\S \otimes \S}^{-1}$ form of the second block of $\mathcal{I}^{-1}_{\bzp}$, specific for the Gaussian loglikelihood. As a consequence, the following three updates are equivalent:
\begin{align}
\iS &\leftarrow R_{\iS}\br{\beta \iS \nabla_{\iS} \iS}\text{,}\nonumber\\
\iS&\leftarrow  R_{\iS}\br{ -\beta \nabla_\S \LB}\text{,}\label{eq:equiv_2}\\
\iS &\leftarrow R_{\iS}\br{\beta \frac{1}{2}\natgrad_{\iS}\LB }\text{.}\label{eq:equiv_3}
\end{align}
All three above correspond to an update of a Riemann gradient on the manifold $\mathcal{M}$ based on the retraction form derived from the manifold $\mathcal{M}$ (specific for the equipped Euclidean metric).
The caveat with \eqref{eq:equiv_3} alone, without Proposition \ref{prop:natgrad} is the following: by removing the factor $\frac{1}{2}$ the argument of the retraction form $\eqref{eq:equiv_3}$ derived from the SPD manifold equipped with the Euclidean metric, is not longer the Riemann gradient for this manifold, but the Riemann gradient of a different manifold (the SPD manifold equipped with the Fisher-Rao metric $\mathcal{F}$). This point is central and discussed in the main text.\\
In support of our argument and derivation, by unfolding the retraction in \eqref{eq:equiv_3}, the update
$$
\iS \leftarrow \iS + \frac{\beta}{2}\sbr{\natgrad_{\iS}\LB} + \frac{\beta^2}{2} \sbr{\natgrad_{\iS}\LB} \S \sbr{\natgrad_{\iS}\LB}\text{,}
$$
aligns with the update form of \citep{lin2020handling} as a special case of their Bayesian learning rule, since the Gaussian FIM satisfies their block-coordinate natural parametrization assumption.

Working on $\S$, one analogously obtains $R_{\S}\br{\S \nabla_\S \LB \S} = R_{\S}\br{\frac{1}{2} \natgrad_\S \LB}$;  however, the latter does not correspond to a simpler algebraic form leading to a computationally convenient update like \eqref{eq:equiv_2}.

\subsection{General form of the MGVBP update}\label{app:proof_MGVBP_cases}
For a prior $p\sim\N\br{\bmu_0,\S_0}$ and a variational posterior $q \sim \N\br{\bmu,\S}$, by rewriting the LB as 
\begin{align*}
  \E_{q_\bz}\sbr{h_\bz\brt} &= \E_{q_\bz}\sbr{\log p\brt - \log \qz \brt + \log \lik}\\
  &= \E_{q_\bz}\sbr{\log p\brt - \log \qz \brt} + \E_{q_\bz}\sbr{\log \lik} \text{,}
\end{align*}
we decompose $\nabla_\bz \LBz$ as $\nabla_\bz \E_{q_\bz}\sbr{\log p\brt - \log \qz \brt} + \nabla_\bz \E_{q_\bz}\sbr{\log p\br{\by\vert \bth}}$. As in \eqref{eq:naive_mc_grad_estimator}, we apply the log-derivative trick on the last term and write $\nabla_\bz \E_{q_\bz} \sbr{\log \lik}=\E_{q_\bz}\sbr{\nabla_\bz\sbr{\log q_\bz\brt} \log p\br{\by\vert \bth}}$. On the other hand, it is easy to show that up to a constant that does not depend on $\bmu$ and $\S$
\begin{align*}
\mathbb{E}_{q_\bz}\sbr{\log p\brt - \log \qz \brt} = &-\frac{1}{2}\log \vert \S_0 \vert +  \frac{1}{2}\log \vert \S \vert  +\frac{1}{2}d \\
&- \frac{1}{2}\text{tr}\br{\S_{0}^{-1} \S} - \frac{1}{2}\br{\bmu - \bmu_0}^\top\S_0^{-1}\br{ \bmu-\bmu_0} \text{,}
\end{align*}
so that
\begin{align*}
    \nabla_\S\E_{q_\bz}\sbr{\log p\brt - \log \qz \brt} &= \frac{1}{2}\iS -\frac{1}{2}\iSz \text{,}\\
    \nabla_\bmu\E_{q_\bz}\sbr{\log p\brt - \log \qz \brt} &= -\iSz\br{\bmu-\bmu_0} \text{.}
\end{align*}
For the natural gradients, we have
\begin{align*}
    \natgrad_{\iS}\E_{q_\bzp}\sbr{\log p\brt - \log q_\bzp \brt}  &= -\nabla_\S\E_{q_\bz}\sbr{\log p\brt - \log \qz \brt} \\&= -\frac{1}{2}\iS + \frac{1}{2}\iSz \text{,}\\
    \natgrad_\bmu\E_{q_\bzp}\sbr{\log p\brt - \log q_\bzp \brt} &= \S \nabla_\bmu\E_{q_\bz}\sbr{\log p\brt - \log \qz \brt} \\&= -\S \iSz\br{\bmu-\bmu_0} \text{,}
\end{align*}
while the naive estimators for $\natgrad_{\bmu} \E_{q_\bzp}\sbr{\log \lik}$ and $\natgrad_{\iS} \E_{q_\bzp}\sbr{\log \lik}$ turn analogous to feasible natural gradient estimators presented in Section \ref{sec:MGVB:implementation}, with $h_\bz$ replaced by $\log p\br{\by\vert \bth}$.
This leads to the general form of the MGVBP update, based either on the $h$-function gradient estimator (generally applicable) or the above decomposition (applicable under a Gaussian prior):
\begin{align*}
      \natgrad_{\bmu} \LB_t \br{\bzp} &\approx  c_{\bmu_t} + \frac{1}{S}\sum_{s=1}^S \sbr{ \br{\bth_s-\bmu_t} \log f\br{\bth_s} }\\
      \natgrad_{\iS} \LB_t\br{\bzp}  &\approx C_{\S_t} + \frac{1}{2S} \sum_{s=1}^S\sbr{\br{\iS_t-\iS_t\br{\bth_s-\bmu_t}\br{\bth_s-\bmu_t}^\top\iS_t} \log f\br{\bth_s} } 
\end{align*}
where 
\begin{equation}\label{eq:cases}
  \begin{cases}
    \begin{cases}
        C_{\S_t} = -\frac{1}{2}\iS_t +\frac{1}{2}\iSz\text{,}\\
        c_{\bmu_t} = -\S_t \iSz\br{\bmu_t-\bmu_0}\text{,}\\
        \log f\br{\bth_s} = \log p\br{\by\vert \bth_s}\text{,}
    \end{cases}
    &\text{if $p$ is Gaussian,}\\\\
    \begin{cases}
        C_{\S_t} = 0\text{,}\\
        c_{\bmu_t} = \bm{0}\text{,}\\  
        \log f\br{\bth_s} = h_{\bz}\br{\bth_s}\text{,}
    \end{cases}
    &\text{in general, whether $p$ is Gaussian or not.}
  \end{cases}
\end{equation}
The short-hand notation $\log f\br{\bth_s}$ embeds the fact that the function is meant to be evaluated at the current value of the parameter, i.e., $\log f\br{\bth_s} \equiv \log f\br{\bth_s\vert \bmu = \bmu_t,\S = \S_t}$.

\end{document}